\documentclass[twoside,11pt]{article}
\usepackage{jmlr2e}
\usepackage[T1]{fontenc}
\usepackage{amsmath, amssymb}%amsthm,
\usepackage{graphicx} % more modern
\usepackage{caption}
\usepackage{subcaption}
\usepackage[noend,linesnumbered]{algorithm2e}
\makeatletter
\renewcommand{\@algocf@capt@plain}{above}% formerly {bottom}
\makeatother

\usepackage{url}
\usepackage{tikz}
\usepackage{color}
\usepackage{multirow}

\usepackage{rotating}
\usepackage{tabu}
\usepackage{euler}
\usepackage[small]{eulervm}
\newcommand{\refSection}[1]{Section~\ref{#1}}
\newcommand{\refTable}[1]{Table~\ref{#1}}
\newcommand{\refFigure}[1]{Figure~\ref{#1}}

\newcommand{\refCitep}[1]{\citeauthor{#1}~\citeyear{#1}}

\newcommand{\ie}[0]{\textit{i.e.},~}
\newcommand{\eg}[0]{\textit{e.g.},~}
\newcommand{\aka}[0]{{a.k.a}.~}

\newcommand{\mb}[0]{\ensuremath{\Delta}}
\newcommand{\x}[0]{\ensuremath{x}}

\newcommand{\y}[0]{\ensuremath{\mathbf{y}}}

\newcommand{\parisi}[0]{\ensuremath{{m}}}
\newcommand{\xh}[0]{\ensuremath{\widehat{x}}}
\newcommand{\set}[1]{\ensuremath{\mathcal{#1}}}
\newcommand{\f}{\ensuremath{C}}

\newcommand{\back}[1]{\ensuremath{\setminus #1}}
\newcommand{\p}{\ensuremath{\mu}}%{\mathbb{P}}}
\newcommand{\mbp}{\ensuremath{{\Lambda}}}
\newcommand{\mbpiI}[2]{\ensuremath{{\Lambda}_{#1 \to #2}}}
\newcommand{\mbpi}[1]{\ensuremath{{\Lambda}_{#1}}}
\newcommand{\bp}{\ensuremath{{\Phi}}}
\newcommand{\bpiI}[2]{\ensuremath{{\Phi}_{#1 \to #2}}}
\newcommand{\pbpiI}[2]{\ensuremath{{\Gamma}_{#1 \to #2}}}
\newcommand{\pbp}{\ensuremath{{\Gamma}}}
\newcommand{\gsi}[1]{\ensuremath{{\Psi}_{#1}}}
\newcommand{\gs}{\ensuremath{{\Psi}}}
\newcommand{\gsspi}[1]{\ensuremath{\boldsymbol{\Psi}_{#1}}}
\newcommand{\gssp}{\ensuremath{\boldsymbol{\Psi}}}
\newcommand{\sppiI}[2]{\ensuremath{\boldsymbol{\Phi}_{#1 \to #2}}}
\newcommand{\spp}{\ensuremath{\boldsymbol{\Phi}}}

\newcommand{\pspp}{\ensuremath{\boldsymbol{\Gamma}}}
\newcommand{\psp}{\ensuremath{\boldsymbol{\mu}}}%\mathscr{P}}}
\newcommand{\psph}{\ensuremath{\widehat{\boldsymbol{\mu}}}}%\mathscr{P}}}}

\newcommand{\ph}{\ensuremath{\widehat{\mu}}}%{\mathbb{P}}}}

\newcommand{\Z}{\ensuremath{Z}}
\newcommand{\Ms}[2]{\ensuremath{\nu}_{#1 \to #2}}%\mathbb{M}_{#1 \to #2}}}
\newcommand{\Msi}[2]{\ensuremath{\eta_{#1 \to #2}}}%\mathbb{M}^{\times}_{#1 \to #2}}}
\newcommand{\Msp}[2]{\ensuremath{\boldsymbol{\nu}_{#1 \to #2}}}%\mathscr{M}_{#1 \to #2}}}
\newcommand{\sumint}{\sum}
\newcommand{\nn}[1]{^{(#1)}}

\ShortHeadings{Perturbed Message Passing for CSP}
{Ravanbakhsh and Greiner}
\firstpageno{1}

\begin{document}
\title{Perturbed Message Passing for Constraint Satisfaction Problems}
%Dismissing Decimation by Perturbed Message Passing for Constraint Satisfaction Problems}

\author{\name Siamak Ravanbakhsh \email mravanba@ualberta.ca \\
       \addr Department of Computing Science\\
       University of Alberta\\
       Edmonton, AB T6E 2E8, CA
       \AND
       \name Russell  Greiner \email rgreiner@ualberta.ca \\
       \addr Department of Computing Science\\
       University of Alberta\\
       Edmonton, AB T6E 2E8, CA}
\editor{Alexander Ihler}

\maketitle
\begin{abstract}
We introduce an efficient message passing scheme for solving Constraint Satisfaction Problems (CSPs), which uses stochastic perturbation of Belief Propagation (BP) and Survey Propagation (SP)
messages to bypass decimation and directly produce a single satisfying assignment.
 Our first CSP solver, called \emph{Perturbed Belief Propagation},
smoothly interpolates two well-known inference procedures;
it starts as BP and ends as a Gibbs sampler,
which produces a single sample from the set of solutions.
Moreover we apply a similar perturbation scheme to SP to produce another
CSP solver, \emph{Perturbed Survey Propagation}.
Experimental results on random and real-world CSPs
show that Perturbed BP is often more successful and at the same time
tens to hundreds of times more efficient than standard BP guided decimation.
Perturbed BP also compares favorably with state-of-the-art
SP-guided decimation, which has a computational complexity that generally scales exponentially
worse than our method (w.r.t. the cardinality of variable domains and constraints).
Furthermore, our experiments with random satisfiability and coloring problems demonstrate
that Perturbed SP can outperform SP-guided decimation, making it the best incomplete
random CSP-solver in difficult regimes.
\end{abstract}


\maketitle


\section{Introduction}
\label{sec:intro}
Probabilistic Graphical Models (PGMs) provide a common ground for
recent convergence of themes in computer science (artificial neural
networks), statistical physics of disordered systems (spin-glasses)
and information theory (error correcting codes).  In particular,
message passing methods have been successfully applied to obtain
state-of-the-art solvers for
Constraint Satisfaction Problems
\citep{mezard_analytic_2002}%; \refCite{ramezanpour2012cavity}; \refCitep{mulet_coloring_2002}).

The PGM formulation of a CSP defines a uniform distribution over the set
of solutions, where each unsatisfying assignment has a zero probability.
In this framework, solving a CSP amounts to producing a sample from
this distribution. To this end, usually an inference procedure
estimates the marginal probabilities, which suggests an assignment to a subset of the most biased
variables. This process of sequentially fixing a subset of variables, called \emph{decimation}, is repeated
until all variables are fixed to produce a solution.  Due to inaccuracy of the marginal estimates,
this procedure gives
an incomplete solver \citep{kautz2009incomplete}, in the sense
that the procedure's failure is not a certificate of unsatisfiability.
An alternative approach is to use message passing to guide a search
procedure that can back-track if a dead-end is reached
\citep[\eg][]{kask2004counting,parisi_backtracking_2033}.
Here using a branch and bound technique and relying on exact solvers,
one may also determine when a CSP is unsatisfiable.

The most common inference procedure for this purpose is Belief
Propagation \citep{pearl_probabilistic_1988}.  However, due to
geometric properties of the set of solutions
\citep{krzakala_gibbs_2007} as well as the complications from the
decimation procedure \citep{coja-oghlan_belief_2010,kroc2009message},
BP-guided decimation fails on difficult
instances. The study of the change in the geometry of solutions has lead
to  Survey Propagation \citep{braunstein_survey_2002} which is a powerful message passing
procedure that is slower than BP (per iteration) but typically remains convergent, even in many situations
when BP fails
to converge.

Using decimation, or other search schemes that are guided by message passing,
usually requires estimating marginals or partition functions, which is
 harder than producing a single solution \citep{valiant1979complexity}.
This paper introduces a message passing scheme to eliminate this requirement,
therefore also avoiding the complications of applying decimation.
 Our alternative has advantage over both BP- and SP-guided
decimation when applied to solve CSPs. Here we consider BP and
Gibbs Sampling (GS) updates as operators -- $\bp$ and $\gs$
respectively -- on a set of messages. We then consider inference
procedures that are convex combination (\ie $\gamma \gs + (1 -
\gamma) \bp$) of these two operators.
Our CSP solver, Perturbed BP, starts at $\gamma = 0$
and ends at $\gamma = 1$, smoothly changing from BP to GS, and finally
producing a sample from the set of solutions.  This change amounts to
stochastic biasing the BP messages towards the current estimate of
marginals, where this random bias increases in each iteration.  This
procedure is often much more efficient than BP-guided decimation
(BP-dec) and sometimes succeeds where BP-dec fails.  Our results on
random CSPs (rCSPs) show that Perturbed BP is competitive with
SP-guided decimation (SP-dec) in solving difficult random instances.

Since SP can be interpreted as BP applied to an ``auxiliary'' PGM
\citep{Braunstein2003}, we can apply the same perturbation scheme
to SP, which we call Perturbed SP.  Note that this system, also, does
not perform decimation and directly produce a solution (without using local search).
Our experiments
show that Perturbed SP is
often more successful than both SP-dec and Perturbed BP in finding
satisfying assignments.

Stochastic variations of BP have been previously proposed to perform
inference in graphical models  \citep[\eg][]{ihler2009particle,noorshams13a}.
However, to our knowledge, Perturbed BP is
the first method to directly combine GS and BP updates.

In the following, \refSection{sec:notation} introduces PGM formulation
of CSP using factor-graph notation.  \refSection{sec:bpdec} reviews the BP equations and
decimation procedure, then \refSection{sec:gs}
casts GS as a message update procedure.  \refSection{sec:pbp}
introduces Perturbed BP as a combination of GS and BP.
\refSection{sec:benchmark} compares BP-dec and Perturbed BP on
benchmark CSP instances, showing that our method is often several
folds faster and more successful in solving CSPs.
\refSection{sec:clustering} overviews the geometric properties of the set of solutions of rCSPs,
then reviews first order Replica Symmetry Breaking
Postulate and the resulting SP equations for CSP.
\refSection{sec:psp} introduces Perturbed SP and
\refSection{sec:results} presents our experimental results for random
satisfiability and random coloring instances close to the unsatisfiability
threshold.  Finally, \refSection{sec:discussion} further discusses the
behavior of decimation and perturbed BP in the light of a geometric
picture of the set of solutions and the experimental results.

\subsection{Factor Graph Representation of CSP}\label{sec:notation}
Let $\x = ( \x_1, \x_2, \ldots, \x_N )$ be a tuple of $N$ discrete
variables $\x_i\in \set{X}_i$, where each $\set{X}_i$ is the domain of
$\x_i$.  Let $I \subseteq \set{N} = \{1,2,\ldots,N\}$ denote a subset
of variable indices and $\x_I \!=\! \{ \x_i\! \mid\! i\in I\}$ be the
(sub)tuple of variables in $\x$ indexed by the subset $I$.  Each
constraint $\f_{I}(\x_{I}):\; \big (\prod_{i\in I} \set{X}_i \big )
\rightarrow \{0,1 \}$ maps an assignment to 1 \emph{iff} that
assignment satisfies that constraint.  Then the normalized product of
all constraints defines a uniform distribution over solutions:
\begin{align}\label{eq:p}
  \p( \x ) \quad \triangleq \quad \frac{1}{\Z} \prod_{ I }
  \f_{I}(\x_{I})
\end{align}
where the partition function $\Z \;= \; \sumint_{\set{X}} \prod_{ I}
\f_{I}(\x_{I})$ is equal to the number of solutions.\footnote{For Eq~\ref{eq:p} to remain valid when the CSP is unsatisfiable, we define $\frac{0}{0} \triangleq 0$.}
Notice that $\p
(\x)$ is non-zero \emph{iff} all of the constraints are satisfied -- that
is $x$ is a solution. With slight abuse of notation we will use probability density
and probability distribution interchangeably.

\begin{example}[$q$-COL:]
  Here, $\x_i \in \set{X}_i = \{1,\ldots,q\}$ is a q-ary variable for
  each $i \in \set{N}$, and we have $M$ constraints; each constraint
  $\f_{i,j}(\x_{i}, \x_j) = 1 - \delta(\x_i, \x_j) $ depends only on
  two variables and is satisfied iff the two variables have different
  values (colors). Here $\delta(x,x')$ is equal to 1 if $x = x'$ and 0 otherwise.

\end{example}

This model can be conveniently represented as a bipartite graph, known
as a {\em factor graph} \citep{kschischang_factor_2001}, which
includes two sets of nodes: variable nodes $\x_i$, and constraint (or
factor) nodes $\f_{I}$.  A variable node $i$ (note that we will often
identify a variable ``$\x_i$'' with its index ``$i$'') is connected to a
constraint node $I$ if and only if $i \in I$. We will use $\partial$
to denote the neighbors of a variable or constraint node in the factor
graph -- that is $\partial I = \{ i \; \mid \; i \in I \}$ (which is
the set $I$) and $\partial i = \{ I \; \mid \; i \in I \}$. Finally we
use $\mb i$ to denote the Markov blanket of node $\x_i$ ($\mb i
= \{ j \in \partial I\; \mid \; I \in \partial i, j \neq i\}$).
% We use the notation $\Ne{i}$ to denote the variables at distance two
% with $\x_i$ in the factor graph -- \ie the set of variables defined
% by $\Ne{i} := \bigcup_{I \ni i} I \back i$.

The marginal of $\p(\cdot)$ for variable $x_i$ is defined as
$$ \p(\x_i) \; \triangleq \;\sumint_{\set{X}_{ \set{N} \back i}} \p(\x)$$
where the summation above is over all variables but $\x_i$.
Below, we use $\ph(\x_i)$ to denote an \emph{estimate} of this marginal.
Finally, we use $\set{S}$ to denote the (possibly empty) set of solutions $\set{S} = \{
\x \in \set{X}\; \mid \; \p(\x) \neq 0\}$.

\begin{figure*}
  \centering
  \begin{subfigure}{.40\textwidth}
    \includegraphics[width=\textwidth]{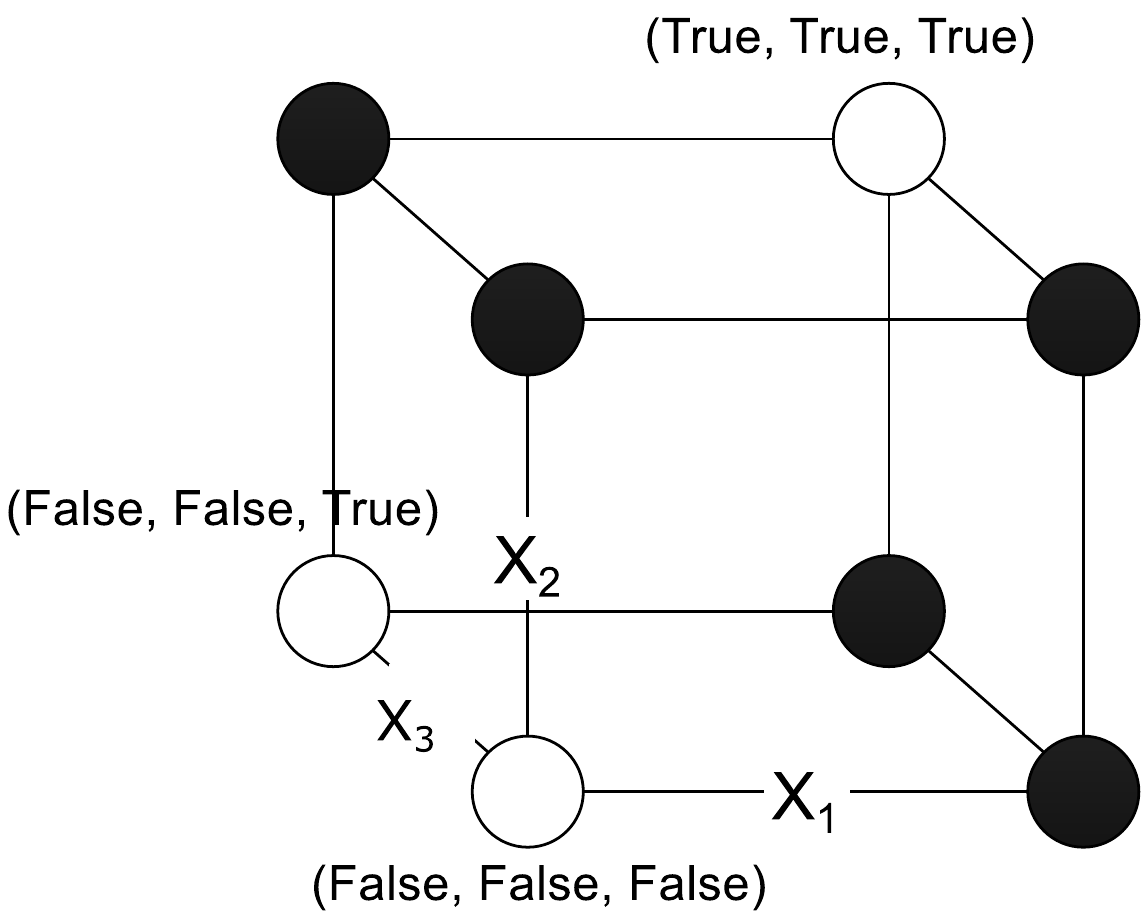}
    \caption{}
    \label{fig:cube}
  \end{subfigure}
  \begin{subfigure}{.40\textwidth}
    \includegraphics[width=\textwidth]{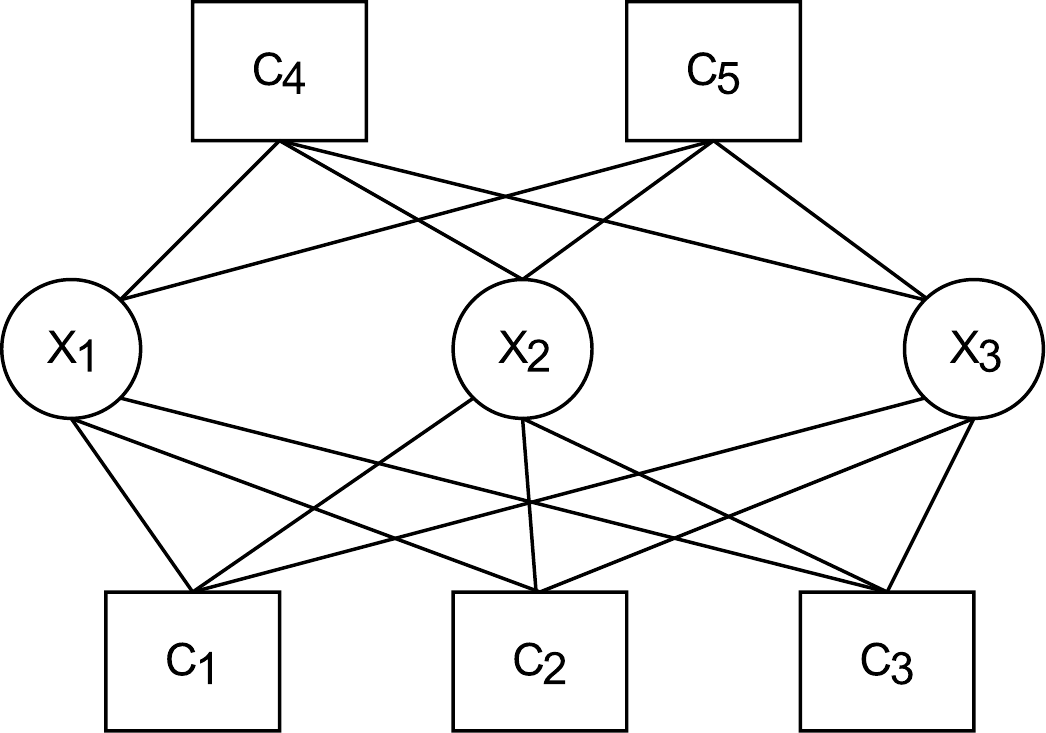}
    \caption{}
    \label{fig:3satfg}
  \end{subfigure}
  \caption{\small{(a) The set of all possible assignments to 3
      variables. The solutions to the 3-SAT problem of
      Eq~\ref{eq:examplesat} are in white
      circles. %Here two neighboring solutions form a cluster, while the third solution has a cluster of its own.
      (b) The factor-graph corresponding to the 3-SAT problem of
      Eq~\ref{eq:examplesat}. Here each factor prohibits a single
      assignment.  }}
  \label{fig:simple3sat}
\end{figure*}

\begin{example}[$\kappa$-SAT:] All variables are binary ($\set{X}_i =
  \{ True,False \}$) and each clause (constraint $\f_I$) depends on $\kappa =
  |\partial I|$ variables. A clause evaluates to 0 only for a single
  assignment out of $2^{\kappa}$ possible assignment of variables
  \citep{garey_computers_1979}.

  Consider the following 3-SAT problem over 3 variables with 5 clauses:
{\small
  \begin{align}\label{eq:examplesat}
    SAT(\x) \; = \; \underbrace{( \neg \x_1 \vee \neg x_2 \vee \x_3)}_{C_1} \wedge \underbrace{( \neg
    \x_1 \vee x_2 \vee \x_3)}_{C_2} \wedge \underbrace{( \x_1 \vee \neg x_2 \vee \x_3)}_{C_3}
    \wedge \underbrace{( \neg \x_1 \vee x_2 \vee \neg \x_3)}_{C_4} \wedge \underbrace{( \x_1 \vee
    \neg x_2 \vee \neg \x_3)}_{C_5}
  \end{align}
}
% \begin{align}\label{eq:examplesat}
%     SAT(\x) \; = \;  ( \neg
%     \x_1 \vee x_2 ) \wedge ( \neg x_2 \vee \x_3)
%     \wedge ( \neg \x_1 \vee x_2 ) \wedge ( \x_1 \vee
%     \neg x_2)
%   \end{align}

  The constraint corresponding to the first clause takes the value
  $1$, except for $x = \{True, True, False \}$, in which case it is
  equal to $0$.
  The set of solutions for this problem is given by:\\
  $\set{S} = \big \{(True, True, True),\; (False, False, False),\;
  ( False, False, True) \big\}$.  \refFigure{fig:simple3sat} shows
  the solutions as well as the corresponding factor graph.\footnote{
    In this simple case, we could combine all the constraints into a
    single constraint over 3 variables and simplify the factor
    graph. However, in general SAT, this cost-saving simplification is
    often not possible.}
  % Here two of solutions form one cluster, while the third solution
  % by itself forms a second cluster.
\end{example}

\subsection{Belief Propagation-guided Decimation}\label{sec:bpdec}
Belief Propagation \citep{pearl_probabilistic_1988} is a
recursive update procedure that sends a sequence of messages from
variables to constraints ($\Ms{i}{{I}}$) and vice-versa
($\Ms{{I}}{i}$):
\begin{align}
  &\Ms{i}{{I}}(\x_i) \;  \propto \; \prod_{J \in \partial i \back I} \Ms{J}{i}(\x_i) \label{eq:miI}\\
  &\Ms{I}{i}(\x_i) \; \propto \; \sumint_{\x_{I \back i} \in \set{X}_{\partial I \back
      i}} \f_{I}(\x_I) \prod_{j \in \partial
    I \back i} \Ms{j}{I}(\x_j) \;
  \label{eq:mIi}
\end{align}
where ${J \in \partial i \back I}$ refers to all the factors connected to variable $\x_i$,
except for factor $\f_I$. Similarly the summation in Eq~\ref{eq:mIi} is over $\set{X}_{\partial I \back i}$, means we are
summing out all $\x_j$ that are connected to $\f_I$ (\ie $x_j\;s.t.\; j \in I \back i$) except for $\x_i$.

The messages are typically initialized to a uniform or a random
distribution.  This recursive update of messages is usually performed
until convergence -- \ie until the maximum change in the value of
all messages, from one iteration to the next, is
negligible (\ie below some small $\epsilon$).
At any point during the updates, the estimated marginal
probabilities are given by
\begin{align}\label{eq:bpmarg}
  \ph(\x_{i}) \; \propto \; \prod_{J \in \partial i} \Ms{J}{i}(\x_i)
\end{align}

In a factor graph without loops, each BP message summarizes the effect
of the (sub-tree that resides on the) sender-side on the receiving
side.

\begin{example}
  Applying BP to the 3-SAT problem of Eq~\ref{eq:examplesat} takes 20
  iterations to converge (\ie for the maximum change in the
  marginals to be below $\epsilon = 10^{-9}$).
Here the message, $\Ms{C_{1}}{1}(x_1)$, from $C1$ to $x_1$ is:
\begin{align*}
\Ms{C_1}{1}(x_1) \quad \propto \quad \sum_{x_{2,3}} C_{1}(x_{1,2,3})\; \Ms{2}{C_1}(x_2)\; \Ms{3}{C_1}(x_3)
\end{align*}

Similarly, the message in the opposite direction, $\Ms{1}{C_1}(x_1)$, is defined as
\begin{align*}
\Ms{1}{C_1}(x_1) \quad \propto \quad \Ms{C_2}{1}(x_1)\; \Ms{C_3}{1}(x_1)\; \Ms{C_4}{1}(x_1)\; \Ms{C_5}{1}(x_1)
\end{align*}

Here BP gives us the
  following approximate marginals: $\ph(x_1 = True) = \ph(\x_2 = True) = .319$
  and $\ph(\x_3 = True) = .522$.  From the set of solutions, we know that
  the correct marginals are $\ph(\x_1 = True) = \ph(\x_2 = True) = 1/3$ and
  $\ph(\x_3 = True) = 2/3$.  The error of BP is caused by influential loops
  in the factor-graph of \refFigure{fig:simple3sat}(b). Here the error
  is rather small; it can be arbitrarily large in some instances or BP
  may not converge at all.
\end{example}
The time complexity of BP updates of Eq~\ref{eq:miI} and Eq~\ref{eq:mIi},
for each of the messages exchanged between $i$ and $I$, are $\set{O}(| \partial i| \ |
\set{X}_i|)$ and
%$\set{O}(|\set{X}_i|^{\ | \partial I|})$
$\set{O}(|\set{X}_I|)$
respectively. We may reduce the time complexity of BP by synchronously updating all the
messages $\Ms{i}{I}\; \forall I \in \partial i$ that leave node $i$. For this, we first
calculate the beliefs $\ph(\x_i)$ using Eq~\ref{eq:bpmarg} and produce each $\Ms{i}{I}$ using
\begin{align}\label{eq:miI_belief}
\Ms{i}{{I}}(\x_i) \;  \propto \; \frac{\ph(\x_i)}{\Ms{I}{{i}}(\x_i)}.
\end{align}
% \vspace{.5in}

Note than we can substitute Eq~\ref{eq:mIi} into Eq~\ref{eq:miI} and
Eq~\ref{eq:bpmarg} and only keep variable-to-factor messages.
After this substitution, BP can be viewed as a fixed-point iteration procedure that repeatedly
applies  the
operator $\bp(\{\Ms{i}{I}\}) \triangleq \{
\bpiI{i}{I}(\{\Ms{j}{J}\}_{j \in \mb i ,J \in \partial i \back I} \})
\}_{i,I \in \partial i}$  to the set of messages
in hope of reaching a fixed point:
\begin{align}
  \Ms{i}{I}(\x_i) \quad \propto \quad \prod_{J \in \partial i \back I}
  \sumint_{\set{X}_{\partial J \back i}} \f_{J}(\x_J) \prod_{j \in \partial J \back i} \Ms{j}{J}(\x_j)
  \quad \triangleq \quad \bpiI{i}{I}(\{\Ms{j}{J}\}_{j \in \mb i,J \in \partial i
    \back I })(\x_i) \label{eq:bpop}
\end{align}
Also Eq~\ref{eq:bpmarg} becomes
\begin{align}\label{eq:bpmargonly}
  \ph(\x_{i}) \quad \propto \quad \prod_{I \in \partial i }
  \sumint_{\set{X}_{\partial I \back i}} \f_{I}(\x_I) \prod_{j \in \partial I \back i} \Ms{j}{I}(\x_j)
\end{align}
where $\bpiI{i}{I}$ denotes individual message
update operators. We let operator $\bp(.)$ denote the set of these $\bpiI{i}{I}$ operators.

\subsubsection{Decimation}
The decimation procedure can employ BP (or SP) to solve a CSP.  We
refer to the corresponding method as BP-dec (or SP-dec).  After
running the inference procedure and obtaining $\ph(\x_i), \; \forall
i$, the decimation procedure uses a heuristic approach to select the most
biased variables (or just a random subset) and fixes these variables
to their most biased values (or a random $\xh_i \sim \ph(\x_i)$). If
it selects a fraction $\rho$ of remaining variables to fix after
each convergence, this multiplies an additional $\log_{\frac{1}{\rho}}(N)$ to the
linear (in $N$) cost\footnote{ Assuming the number of edges in the
  factor graph are in the order of $N$.
In general, using synchronous update of
 Eq~\ref{eq:miI_belief} and assuming a constant factor cardinality, $|\partial I|$, the cost of each
  iteration is $\set{O}(E)$, where $E$ is the number of edges in the
  factor-graph.
} for each iteration of BP (or SP).  The following
algorithm~\ref{alg:bpdec} summarizes BP-dec with a particular scheduling of updates:

\begin{algorithm}[h!]
\caption{Belief Propagation-guided Decimation (BP-dec)}\label{alg:bpdec}
\SetKwInOut{Input}{input}\SetKwInOut{Output}{output}
\SetKwInOut{return}{return}
\DontPrintSemicolon
 \Input{factor-graph of a CSP}
 \Output{a satisfying assignment $x^*$ if an assignment
  was found. \emph{\textsc{unsatisfied}} otherwise}
\BlankLine
\DontPrintSemicolon
initialize the messages\;
$\widetilde{\set{N}} \leftarrow \set{N}$ (set of all variable
  indices) \;
 \While(\tcp{decimation loop}){$\widetilde{\set{N}}$ is not empty}{
 \Repeat(\tcp{BP loop}){convergence}{
\ForEach{$i \in \widetilde{\set{N}}$}{
calculate messages $\{\Ms{I}{i}\}_{I \in \partial i}$ using Eq~\ref{eq:mIi} \;
\If{$\{ \Ms{I}{i} \}_{I \in \partial i}$ are contradictory}{\return{\textsc{unsatisfied}}}
\nllabel{step:deccontradictory}
%\label{step:deccontradictory}
calculate marginal $\ph(\x_i)$ using Eq~\ref{eq:bpmarg} \;
calculate messages $\{\Ms{i}{I}\}_{I \in \partial i}$ using Eq~\ref{eq:miI} or Eq~\ref{eq:miI_belief}\;
 }
}
select $\set{B} \subseteq{\widetilde{\set{N}}}$ using
  $\{\ph(\x_i)\}_{i \in \widetilde{\set{N}}}$\;
fix $ \x^*_j \leftarrow \arg_{\x_j} \max \ph(x_j) \quad \forall j \in \set{B}$\;
reduce the constraints $\{\f_I\}_{I \in \partial j}$ for every $j \in \set{B}$ \;
}
\return{$\x^* = ( x^*_1,\ldots,\x^*_N) $}
\BlankLine
\end{algorithm}

The condition of line \ref{step:deccontradictory} is satisfied iff the product of incoming
messages to node $i$ is $0$ for all $\x_i \in \set{X}_i$.  This means
that neighboring constraints have strict disagreement about the value
of $x_i$ and the decimation has found a contradiction. This
contradiction can happen because, either (I) there is no solution for
the reduced problem even if the original problem had a solution, or
(II) the reduced problem has a solution but the BP messages are
inaccurate.

\begin{example}
  To apply BP-dec to previous example, we first calculate BP
  marginals, as shown in the example above.  Here $\ph(x_1)$ and $\ph(x_2)$
  have the highest bias. By fixing the value of $x_1$ to $False$, the
  SAT problem of Eq~\ref{eq:examplesat} collapses to:
  \begin{align}
    SAT(\x_{\{2,3\}})|_{x_1 = False} \; = \; ( \neg x_2 \vee \x_3)
    \wedge ( \neg x_2 \vee \neg \x_3)
  \end{align}
  BP-dec applies BP again to this reduced problem, which give
  $\ph(\x_2 = True) = .14$ (note here that $\p(x_2 = True) = 0$) and
  $\ph(\x_3 = True) = 1/2$. By fixing $\x_2$ to $False$, another round of
  decimation yields a solution $\x^* = (False, False, True )$.
\end{example}

\vspace{.5in}

\subsection{Gibbs Sampling as Message Update}\label{sec:gs}
Gibbs Sampling (GS) is a Markov Chain Monte Carlo (MCMC) inference
procedure \citep{Andrieu2003} that can produce a set of samples
$\xh[1], \ldots, \xh[L]$ from a given PGM. We can then recover the
marginal probabilities, as empirical expectations:
\begin{align}\label{eq:empavg}
  \ph^{L}(\x_i) \quad \propto \quad \frac{1}{L} \sum_{n=1}^{L}
  \delta(\xh[{n}]_i,\x_i)
\end{align}

Our algorithm only considers a single particle $\xh = \xh[1]$.  GS starts
from a random initial state $\xh\nn{t = 0}$ and at each time-step $t$,
updates each $\xh_i$ by sampling from:
\begin{align}\label{eq:gs}
  \xh\nn{t}_i \; \sim \; \p(\x_i) \quad \propto \quad \prod_{I
    \in \partial i} \f_{I}(\x_i , \xh^{(t-1)}_{\partial I \back i})
\end{align}

If the Markov chain satisfies certain basic properties
\citep{casella1999monte}, $\x_i\nn{\infty}$ is guaranteed to be
an unbiased sample from $\p(\x_i)$ and therefore our marginal
estimate, $\ph^L(\x_i)$, becomes exact as $L \to \infty$.

In order to interpolate between BP and GS, we establish a
correspondence between a particle in GS and a set of variable-to-factor
messages -- \ie $\xh \Leftrightarrow \{\Ms{i}{I}(.)\}_{i, I
  \in \partial i}$.  Here all the messages leaving variable $\x_i$ are
equal to a $\delta$-function defined based on $\xh_i$:
\begin{align}
  \Ms{i}{I}(\x_i) \; = \; \delta(\x_i, \xh_i) \quad \forall I
  \in \partial i
\end{align}

We define the random GS operator $\gs = \{ \gsi{i}\}_i$
and rewrite the GS update of Eq~\ref{eq:gs} as
\begin{align}\label{eq:gsop}
  \Ms{i}{I}(\x_i) \quad \triangleq \quad  \gsi{i}(\{\Ms{j}{J}(\x_j) \}_{j \in
    \mb i, J \in \partial i})(\x_i) \quad = \quad \delta(\xh_i, x_i)
\end{align}
where $\xh_i$ is sampled from
\begin{align}
  \xh_i \; \sim \; \ph(\x_i)  \quad &\propto \quad \prod_{J \in \partial i}
  \f_{I}(\x_i , \xh_{\partial I \back i})\notag\\
  & \propto \quad \prod_{I \in \partial i } \sumint_{\set{X}_{\partial I
      \back i}} \f_{I}(\x_I) \prod_{j \in \partial I \back i}
  \Ms{j}{I}(\x_j)\label{eq:gsnbp}
\end{align}

Note that Eq~\ref{eq:gsnbp} is identical to BP
estimate of the marginal Eq~\ref{eq:bpmargonly}.  This equality is a
consequence of the way we have defined messages in the GS update and
allows us to combine BP and GS updates in the following section.

\section{Perturbed Belief Propagation}\label{sec:pbp}
Here we introduce an alternative to decimation that does not require repeated application of inference.
The basic idea is to use a linear combination of BP and GS operators (Eq~\ref{eq:bpop} and Eq~\ref{eq:gsop}) to update the messages:
\begin{align}
\pbp(\{\Ms{i}{I}\}) \triangleq \gamma \; \gs(\{\Ms{i}{I}\}) + (1 - \gamma) \bp(\{\Ms{i}{I}\})
\end{align}

The Perturbed BP operator $\pbp = \{ \pbpiI{i}{I}\}_{i, I \in \partial i}$ updates each message
by calculating the outgoing message according to BP and GS operators and linearly combines
them to get the final message. During $T$ iterations of Perturbed BP,
the parameter $\gamma$ is gradually and linearly
changed from $0$ towards $1$. Algorithm 2 below summarizes this procedure.

\begin{algorithm}[h!]
\caption{Perturbed Belief Propagation}\label{alg:pbp}
\SetKwInOut{Input}{input}\SetKwInOut{Output}{output}
\SetKwInOut{return}{return}
\DontPrintSemicolon
 \Input{factor graph of a CSP, number of iterations T}
 \Output{a satisfying assignment $x^*$ if an assignment
  was found. \emph{\textsc{unsatisfied}} otherwise}
\BlankLine
\DontPrintSemicolon
initialize the messages\;
$\gamma \leftarrow 0$\;
$\widetilde{\set{N}} \leftarrow \set{N}$ (set of all variable
  indices) \;
%\Repeat{$T$ times }{
\For{$t = 1$ to $T$}{
\ForEach{ variable $x_i$}{
calculate $\Ms{I}{i}$ using Eq~\ref{eq:mIi} $\forall I \in \partial i$ \;
\If{$\{ \Ms{I}{i} \}_{I \in \partial i}$ are contradictory}{\return{\textsc{unsatisfied}}}
\nllabel{step:contradictory}
calculate marginals $\ph(\x_i)$ using Eq~\ref{eq:gsnbp} \nllabel{step:calcmarg} \;
calculate BP messages $\Ms{i}{I}$ using Eq~\ref{eq:miI} or Eq~\ref{eq:miI_belief} \quad $\forall
    I \in \partial i$.\nllabel{step:calcmessage} \;
sample $\xh_i \sim \ph(\x_i)$ \nllabel{step:sample} \;
combine BP and Gibbs sampling messages:
    $\Ms{i}{I} \ \leftarrow \  \gamma \  \Ms{i}{I}\ \ + \  (1 - \gamma)\ \delta(\x_i, \xh_i)$\nllabel{step:projection} \;

 }
$\gamma \leftarrow \gamma + \frac{1}{T-1}$\;
}
\return{$\x^* = \{ x^*_1,\ldots,\x^*_N\} $}
\BlankLine

\end{algorithm}

In step \ref{step:contradictory}, if the product of incoming messages is $0$ for all $\x_i \in \set{X}_i$ for some $i$,
different neighboring constraints have strict disagreement about
$\x_i$; therefore  this run of Perturbed BP will not be able to satisfy this CSP.
Since the procedure is inherently stochastic, if the CSP is satisfiable,
re-application of the same procedure to the problem may avoid
this specific contradiction.

\subsection{Experimental Results on Benchmark
  CSP}\label{sec:benchmark}
This section compares the performance of BP-dec and Perturbed BP on
benchmark CSPs. We considered CSP instances from XCSP repository
\citep{roussel2009xml,xcsp}, without global
constraints or complex domains.\footnote{All instances with intensive
constraints (\ie functional form) were converted into extensive format
for explicit representation using dense factors.  We further removed
instances containing constraints with more that $10^6$ enteries in
their tabular form.  We also discarded instances that collectively had
more than $10^8$ enteries in the dense tabular form of their
constraints. Since our implementation represents all factors
  in a dense tabular form, we had to remove many instances
  because of their large factor size. We anticipate that
Perturbed BP and BP-dec could probably solve many of these instances
using a sparse representation.}

We used a convergence threshold of $\epsilon = .001$ for BP and
terminated if the threshold was not reached after $T = 10
\times 2^{10} = 10,240$ iterations. To perform decimation, we sort the
variables according to their bias and fix $\rho$ fraction of the most biased
variables in each iteration of decimation. This fraction, $\rho$,
was initially set to $100\%$, and it was divided by $2$ each time
BP-dec failed on the same instance. BP-dec was repeatedly applied
using the reduced $\rho$, at most $10$ times, unless a solution was
reached (that is $\rho = .1\%$ at final attempt).

For Perturbed BP, we set $T = 10$ at the starting attempt, which was
increased by a factor of $2$ in case of failure. This was repeated at
most $10$ times, which means Perturbed BP used $T = 10,240$ at its final
attempt. Note that Perturbed BP at most uses the same number of
iterations as the maximum iterations per single iteration of decimation in
BP-dec.
\begin{figure}
  \centering
\hbox{
  \begin{subfigure}{.45\textwidth}
    \includegraphics[width=\textwidth]{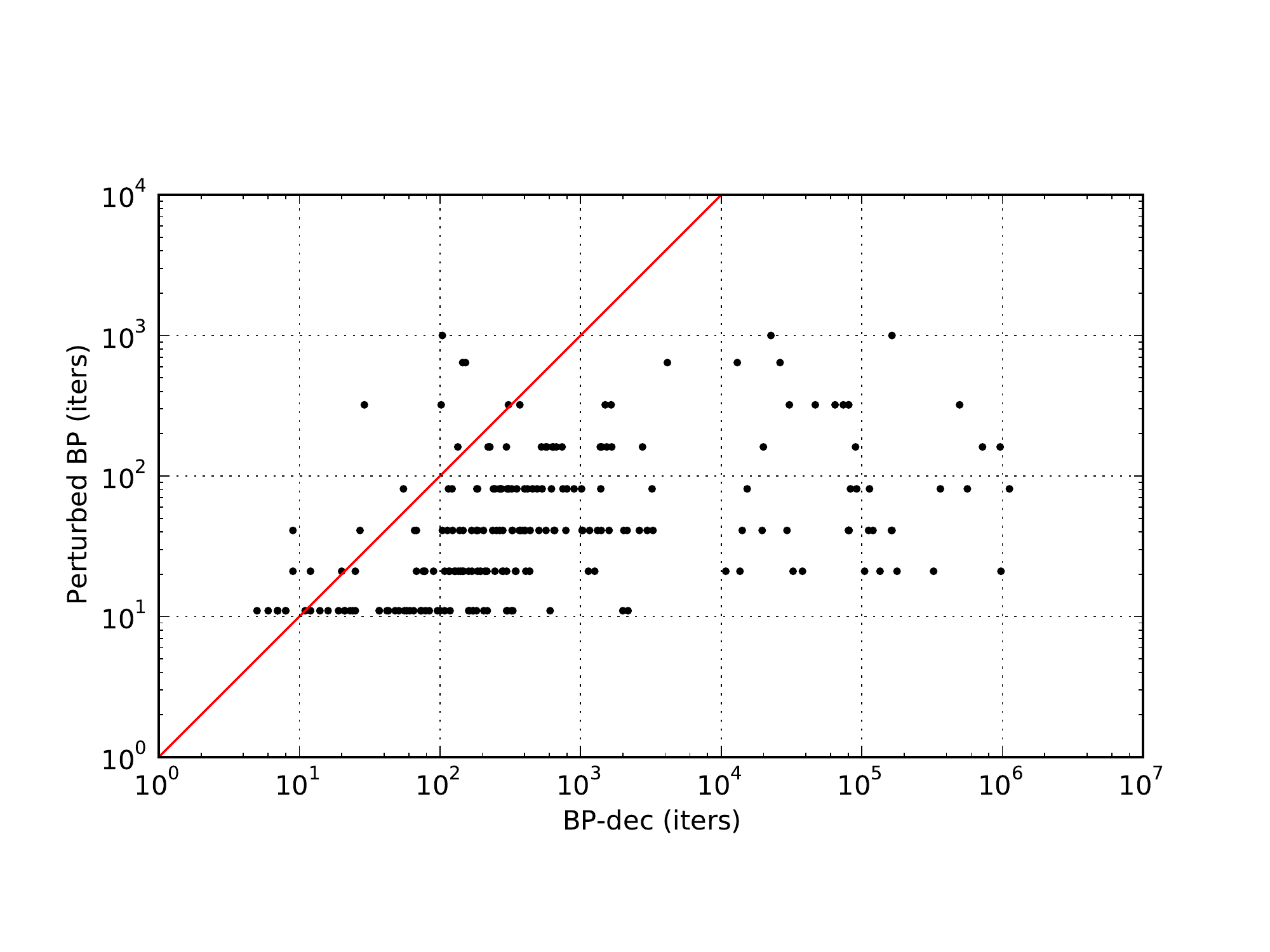}
    \caption{iterations}
  \end{subfigure}
  \begin{subfigure}{.45\textwidth}
    \includegraphics[width=\textwidth]{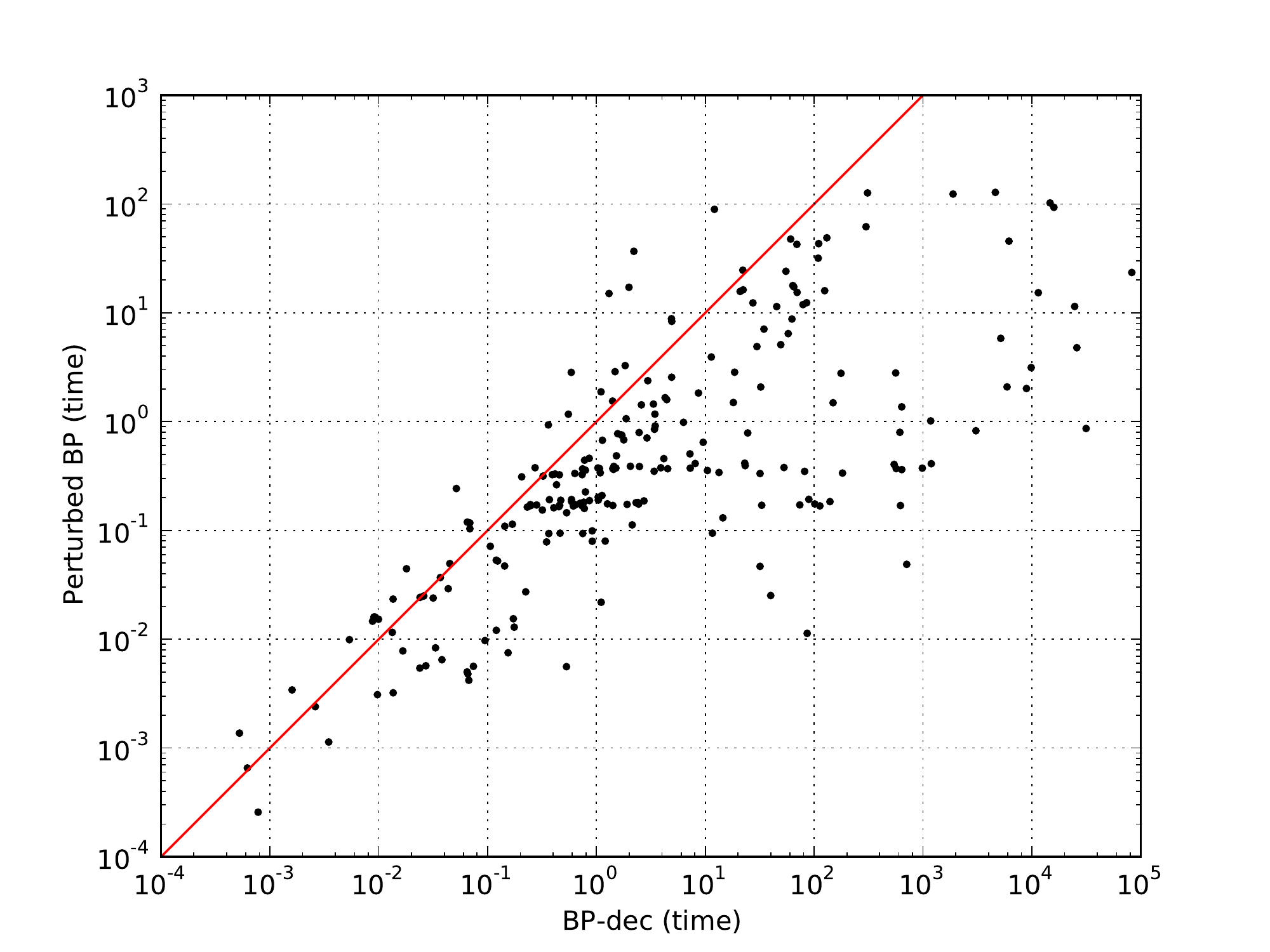}
    \caption{time (seconds)}
\end{subfigure}
}
\hbox{
  \begin{subfigure}{.45\textwidth}
    \includegraphics[width=\textwidth]{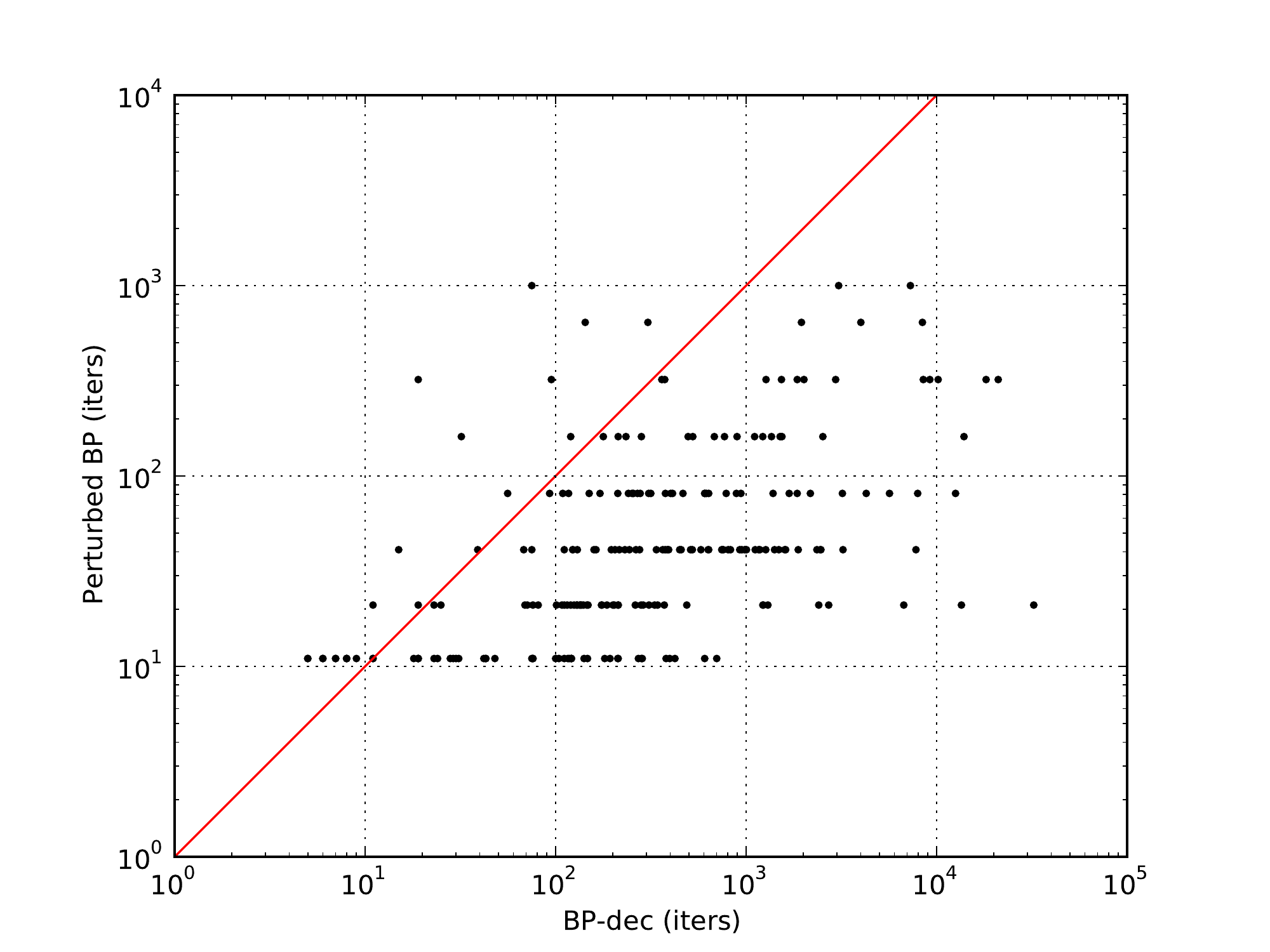}
    \caption{iterations for $T = 1000$}
  \end{subfigure}
  \begin{subfigure}{.45\textwidth}
    \includegraphics[width=\textwidth]{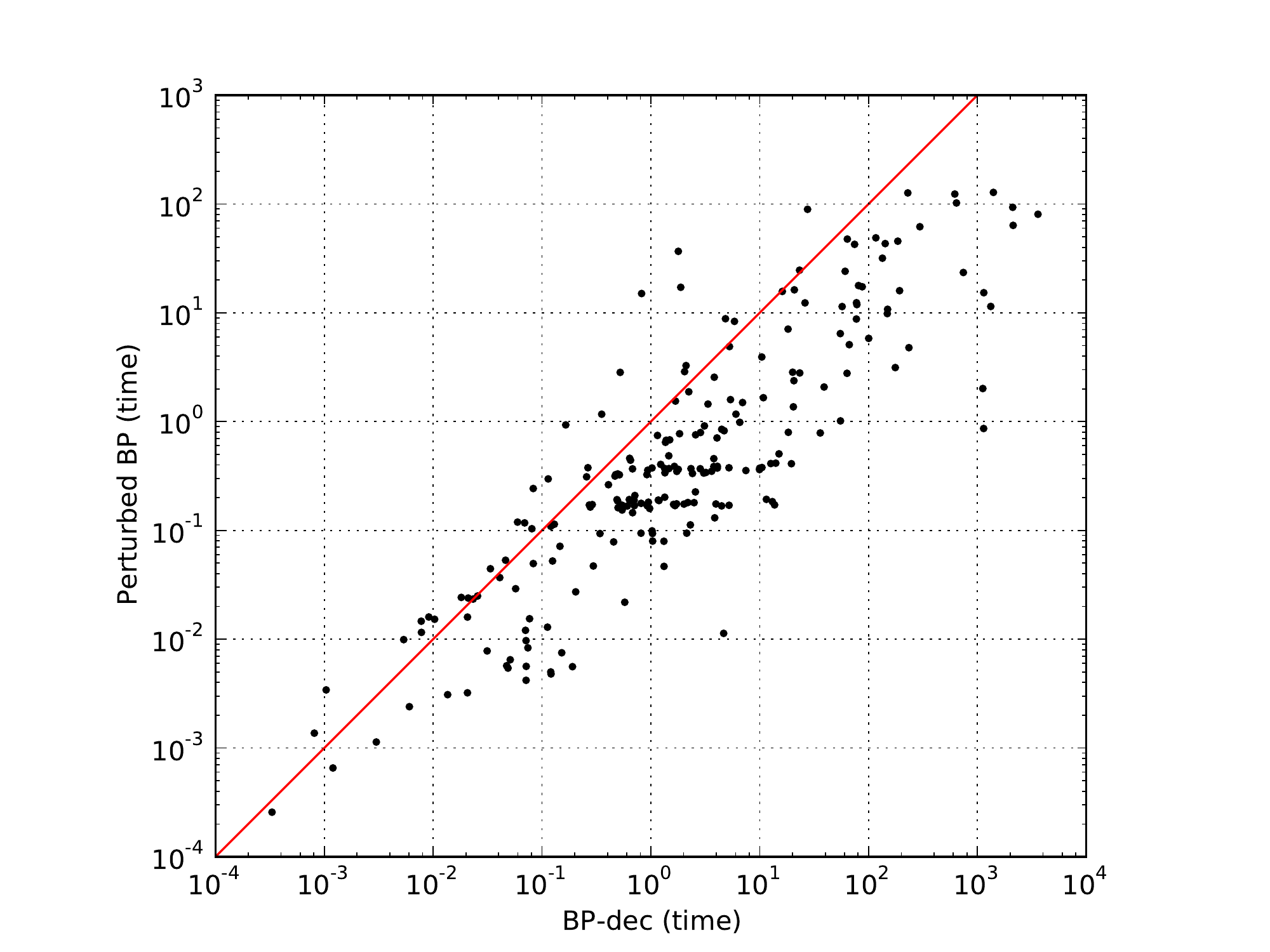}
    \caption{time (seconds) for $T = 1000$}
  \end{subfigure}
}
\hbox{
  \begin{subfigure}{.45\textwidth}
    \includegraphics[width=\textwidth]{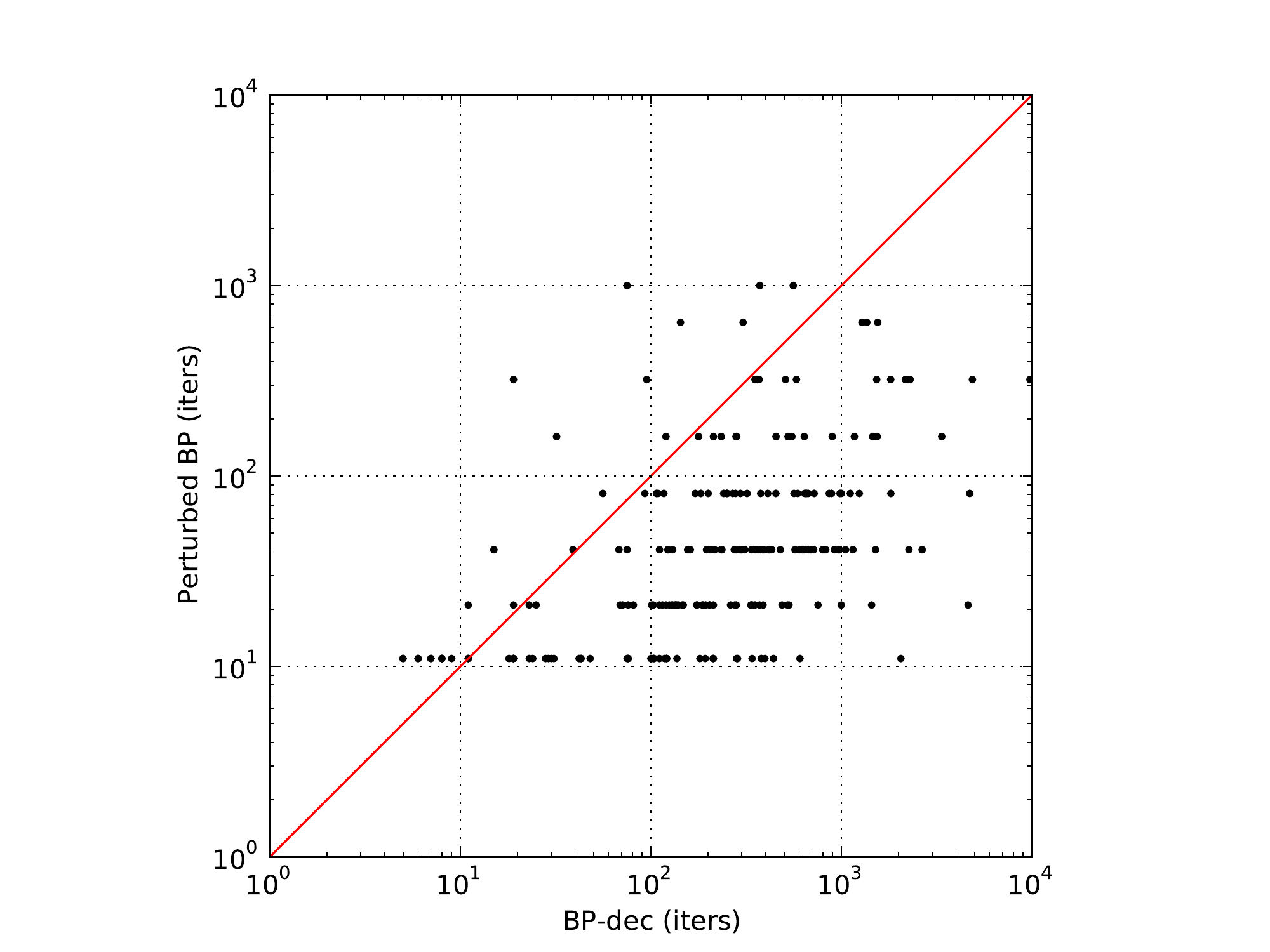}
    \caption{iterations for $T = 100$}
  \end{subfigure}
  \begin{subfigure}{.45\textwidth}
    \includegraphics[width=\textwidth]{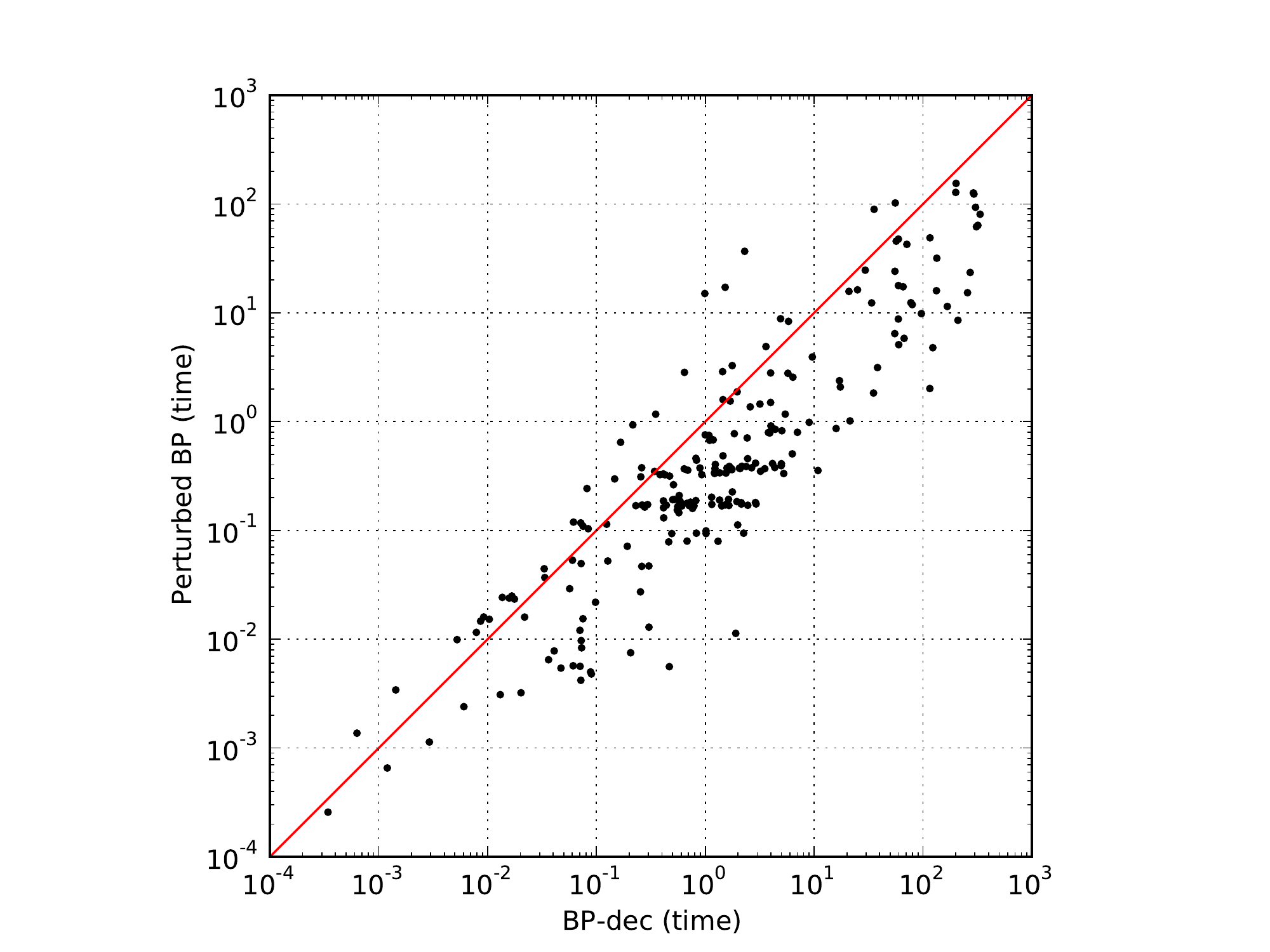}
    \caption{time (seconds) for $T = 100$}
  \end{subfigure}
}
  \caption{\small{Comparison of time and number of iterations used by BP-dec and Perturbed BP
in benchmark instances where both methods found satisfying assignments. (a,b) Maximum number of BP iterations
per iteration of decimation is $T = 10,240$, equal to the maximum iterations used by Perturbed BP.
 (c,d) Maximum number of iterations for BP in BP-dec is reduced to $T = 1000$.
(e,f) Maximum number of iterations for BP in BP-dec is further reduced to $T = 100$.}}
  \label{fig:benchmark}
\end{figure}

\refFigure{fig:benchmark}(a,b) compares the time and iterations of BP-dec and
Perturbed BP for successful attempts where both methods satisfied an instance.
The result for individual problem-sets is reported in the appendix.

Empirically, we found that Perturbed BP both solved (slightly) more instances than BP-dec (284 vs 253), and was (hundreds of times) more efficient: while Perturbed BP  required only 133 iterations on average, BP-dec required an average of 41,284 iterations for successful instances.

We also ran BP-dec on all the benchmarks with maximum number of
iterations set to $T = 1000$ and $T = 100$ iterations.  This reduced
the number of satisfied instances to $249$ for $T = 1000$ and $247$
for $T = 100$, but also reduced the average number of iterations to
$1570$ and $562$ respectively, which are still several folds more
expensive than Perturbed BP. \refFigure{fig:benchmark}(c-f) compare the
time and iterations used by BP-dec in these settings with that of Perturbed BP,
when both methods found a satisfying assignment.
See the appendix for a more detailed report on these results.

\section{Critical Phenomena in Random CSPs}\label{sec:clustering}

Random CSP (rCSP) instances have been extensively used in order to
study the properties of combinatorial problems
\citep{mitchell1992hard,achioptas2000optimal,krzakala_gibbs_2007}
as well as in analysis and design of algorithms
\citep[\eg][]{selman1994noise,mezard_analytic_2002}.
Random CSPs are closely related to spin-glasses in statistical physics
\citep{kirkpatrick1994critical,fu_application_1986}.
This connection follows from the fact that the Hamiltonian of these
spin-glass systems resembles the objective functions in many
combinatorial problems, which decompose to pairwise (or higher
order) interactions, allowing for a graphical representation in the
form of a PGM. Here message passing methods, such as belief
propagation (BP) and survey propagation (SP), provide consistency
conditions on locally tree-like neighborhoods of the graph.

The analogy between a physical system and computational problem
extends to their critical behavior where computation relates to
dynamics \citep{ricci-tersenghi_being_2010}.  In computer
science, this critical behavior is related to the time-complexity of
algorithms employed to solve such problems, while in spin-glass
theory this translates to dynamics of glassy state, and exponential
relaxation times \citep{Mezard1987}.  In fact, this connection
has been used to attempt to prove the conjecture that $\set{P}$ is not
equal to $\set{NP}$ \citep{deolalikar}.

Studies of rCSP, as a critical phenomena, focus on the geometry of the
solution space as a function of the problem's difficulty, where
rigorous
\citep[\eg][]{achlioptas_algorithmic_2008,cocco_rigorous_2002}
and non-rigorous  (\eg Cavity method of
\citet{mezard_bethe_2000} and \citet{mezard_cavity_2002}) analyses
have confirmed the same geometric picture.

When working with large random instances, a scalar $\alpha$ associated
with a problem instance, \aka control parameter -- for example, the clause to
variable ratio in SAT -- can characterize that instance's difficulty
(\ie larger control parameter corresponds to a more difficult
instance) and in many situations it characterizes a sharp transition
from satisfiability to unsatisfiability
\citep{cheeseman_where_1991}.

\begin{example}[Random $\kappa$-SAT]
  Random $\kappa$-SAT instance with $N$ variables and $M = \alpha N$
  constraints are generated by selecting $\kappa$ variables at random
   for each constraint.  Each constraint is set to
  zero (\ie unsatisfied) for a single random assignment (out of
  $2^{\kappa}$).  Here $\alpha$ is the control parameter.
\end{example}
\begin{example}[Random $q$-COL]
  The control parameter for a random $q$-COL instances with $N$
  variables and $M$ constraints is its average degree $\alpha =
  \frac{2M}{N}$.  We consider Erd\H{o}s-R\'{e}ny random graphs and
  generate a random instance by sequentially selecting two distinct variables out of $N$ at
  random to generate each of $M$ edges. For large $N$, this is
  equivalent to selecting each possible factor with a fixed
  probability, which means the nodes have Poisson degree distribution
  $\mathbb{P}(|\partial i| = d) \propto e^{-\alpha} \alpha^{d}$.
\end{example}

\begin{figure}
  \centering
  \begin{subfigure}{.3\textwidth}
    \includegraphics[width=\textwidth]{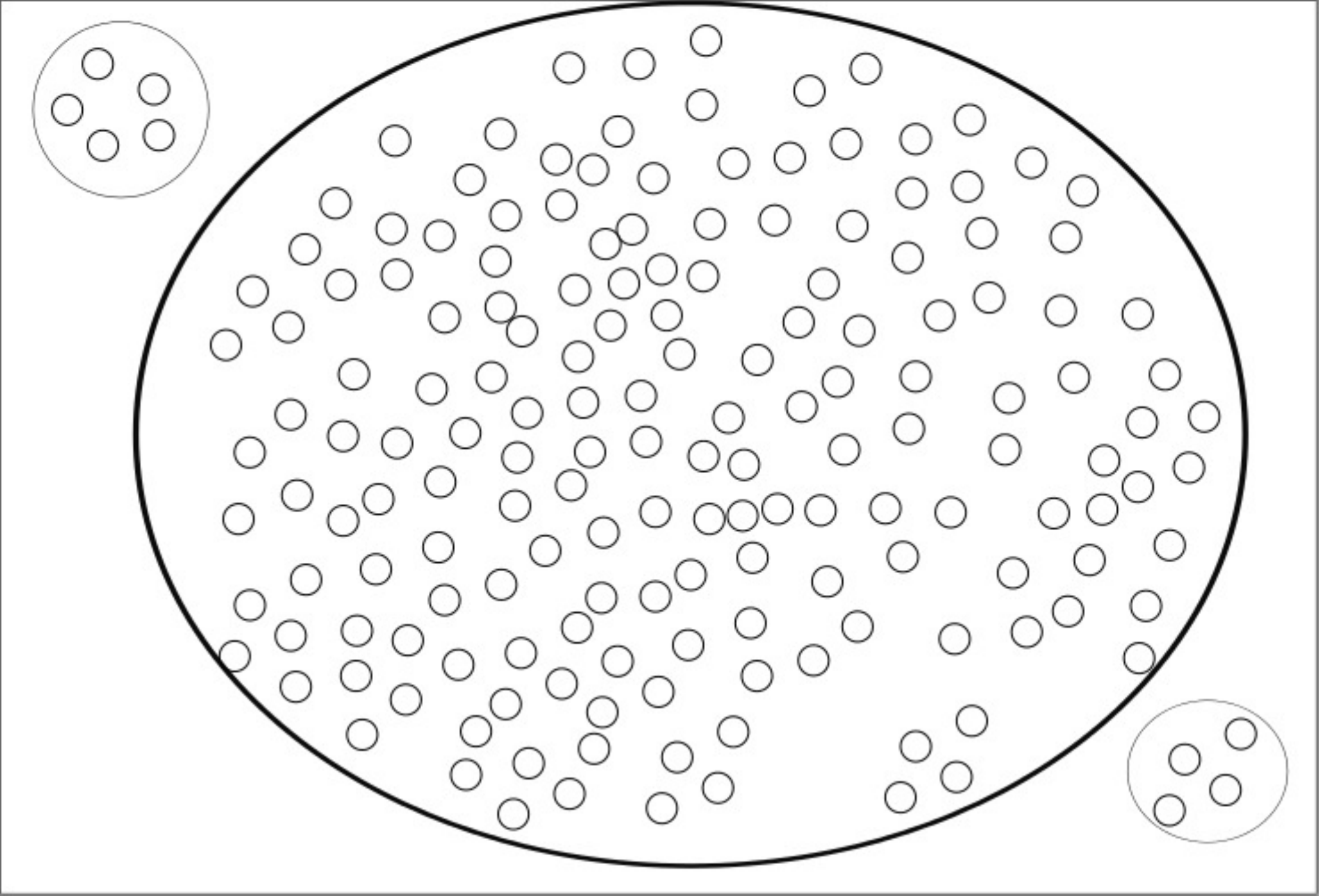}
    \caption{Replica Symmetric}
    \label{fig:RS}
  \end{subfigure}
  \begin{subfigure}{.3\textwidth}
    \includegraphics[width=\textwidth]{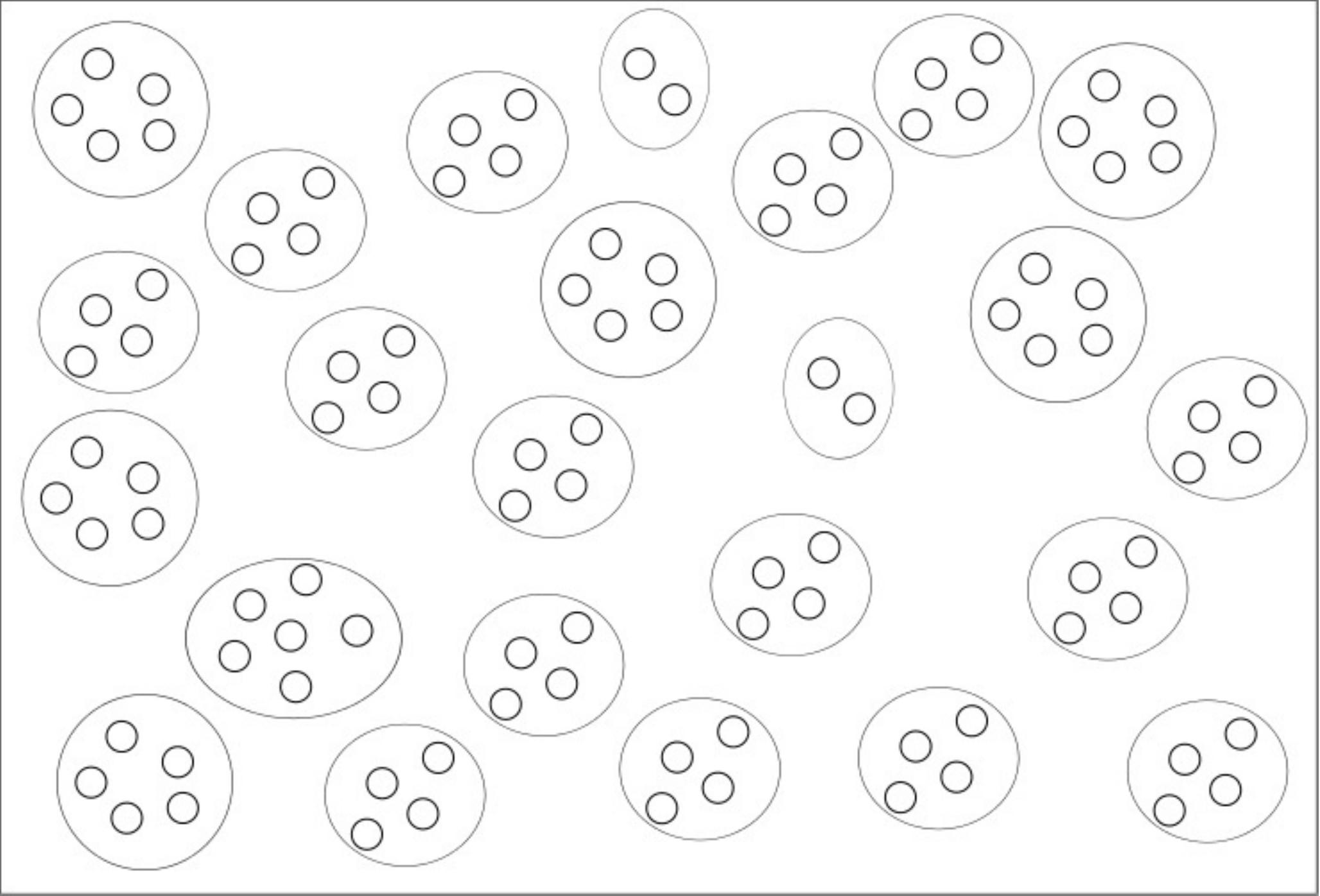}
    \caption{clustering}
    \label{fig:dynamical}
  \end{subfigure}
  \begin{subfigure}{.3\textwidth}
    \includegraphics[width=\textwidth]{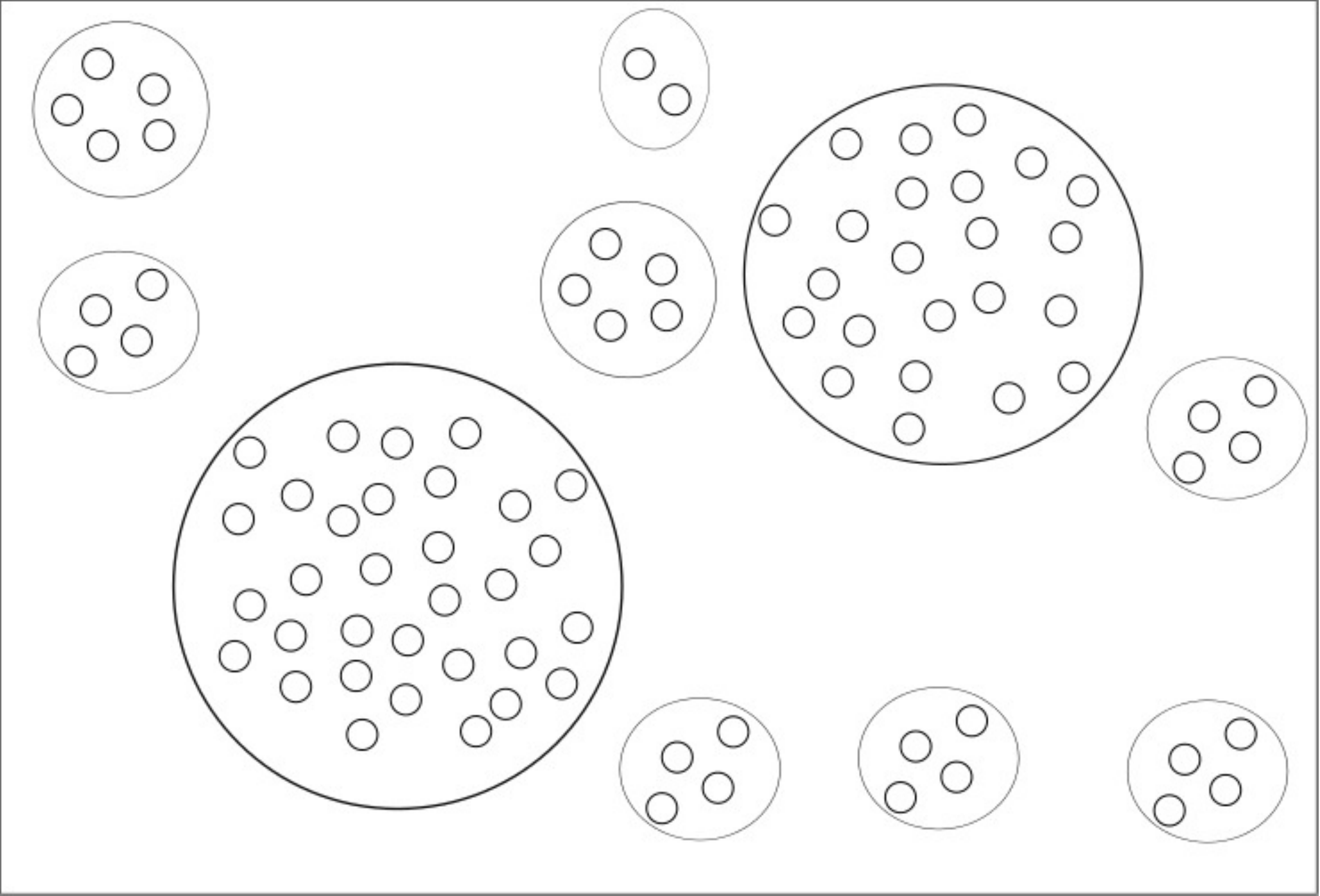}
    \caption{condensation}
    \label{fig:condensation}
  \end{subfigure}
  % \begin{subfigure}{.23\textwidth}
  %   \includegraphics[width=\textwidth]{figures/perturbedbp}
  %   \caption{Perturbed BP}
  %   \label{fig:perturbedbp}
  % \end{subfigure}
  \caption{\small{ A 2-dimensional schematic view of how the set of
      solutions of CSP varies as we increase the control parameter
      $\alpha$ from (a)~replica symmetric phase to (b)~clustering
      phase to (c)~condensation phase.  Here small circles represent
      solutions and the bigger circles represent clusters of
      solutions.   Note that this view is very simplistic in many ways;
      for example, the total number of solutions and the size of clusters
      should generally decrease from (a) to (c).}  }
% (d)~the rightmost figure demonstrates how our
%       method, Perturbed BP, operates.  The big light-grey ellipse
%       represent BP's distribution over the set of solutions.
%       Perturbed BP iteratively converges towards a particular cluster
%       and finally a particular solution (shown by darker, and smaller,
%       ellipses).
  \label{fig:schematic}
\end{figure}

While there are tight bounds
for some problems \citep[\eg][]{achlioptas_rigorous_2005}, finding
the exact location of this transition for different CSPs is still an
open problem.  Besides transition to unsatisfiability, these analyses
has revealed several other (phase) transitions
\citep{krzakala_gibbs_2007}.
\refFigure{fig:schematic}(a)-(c)
shows how the geometry of the set of solutions changes by increasing
the control parameter.

Here we enumerate various phases of the problem for increasing values
of the control parameter: \textbf{(a)} In the so-called \emph{Replica
  Symmetric} (RS) phase, the symmetries of the set of solutions (\aka
ground states) reflect the trivial symmetries of problem w.r.t.~variable
domains. For example, for $q$-COL the set of solutions is symmetric
w.r.t.~swapping all red and blue assignment.  In this regime, the set of
solutions form a giant cluster (\ie a set of neighboring solutions),
where two solutions are considered neighbors when their Hamming
distance is one \citep{achlioptas_algorithmic_2008} or
non-divergent with number of variables
\citep{mezard_cavity_2002}.  Local search methods \citep[\eg][]{selman1994noise} and BP-dec can often
efficiently solve random CSPs that belong to this phase.

\textbf{(b)} In \emph{clustering} or \emph{dynamical} transition
(1dRSB\footnote{1st order dynamical
  RSB. % Symmetry Breaking is a general term indicating a phenomenon
  % during which a system is breaking the symmetry that governs its
  % behaviour by selecting a particular branch.
  The term Replica Symmetry Breaking (RSB) originates from the
  technique -- \ie Replica trick (\refCitep{Mezard1987}) -- that was
  first used to analyze this setting. According to RSB, the trivial
  symmetries of the problem do not characterize the clusters of
  solution.}), the set of solutions decomposes into an exponential
number of distant clusters.  Here two clusters are distant if the
Hamming distance between their respective members is divergent (\eg
linear) in the number of variables.  \textbf{(c)} In the \emph{condensation}
phase transition (1sRSB\footnote{1st order static RSB.}), the set of
solutions condenses into a few dominant clusters. Dominant clusters
have roughly the same number of solutions and they collectively
contain almost all of the solutions.  While SP can be used even within
the condensation phase, BP usually fails to converge in this regime.
However each cluster of solutions in the clustering and condensation
phase is a valid fixed-point of BP, which is called a
``quasi-solution'' of BP.  \textbf{(d)} A \emph{rigidity} transition (not
included in \refFigure{fig:schematic}) identifies a phase in which a
finite portion of variables are fixed within dominant clusters.  This
transition triggers an exponential decrease in the total number of
solutions, which leads to \textbf{(e)} unsatisfiability transition.\footnote{In
  some problems, the rigidity transition occurs before condensation
  transition.}
This rough picture summarizes first order Replica
Symmetry Breaking's (1RSB) basic assumptions \citep{Mezard2009}.

From a geometric perspective, the intuitive idea behind Perturbed BP, is
to perturb the messages towards a solution.  However, in order
to achieve this, we need to initialize the messages to a proper
neighborhood of a solution.  Since these neighborhoods are not
initially known, we resort to stochastic perturbation of messages to
make local marginals more biased towards a subspace of solutions.
This continuous perturbation of all messages is performed in a way
that allows each BP message to re-adjust itself to the other
perturbations, more and more focusing on a random subset of solutions.

\subsection{1RSB Postulate and Survey Propagation} \label{sec:1rsb}
Large random graphs are locally tree-like, which means the length of short
loops are typically in the order of $\log(N)$
\citep{janson_random_2001}.  This ensures that, in the absence of
long-range correlations, BP is asymptotically exact, as the set of
messages incoming to each node or factor are almost independent.
Although BP messages remain uncorrelated until the condensation
transition \citep{krzakala_gibbs_2007}, the BP equations do not
completely characterize the set of solutions after the clustering
transition.  This inadequacy is indicated by the existence of a set of
several valid fixed points (rather than a unique fixed-point) for BP, each of
which corresponds to a quasi-solution.  For a better intuition,
consider the cartoons of Figures~\ref{fig:schematic}(b) and
(c). During the clustering phase (b), $x_i$ and $x_j$ (corresponding to the $x$ and $y$ axes)
are not highly
correlated, but they become correlated during and after condensation
(c). This correlation between variables that are far apart in the PGM
results in correlation between the BP messages. This violates
 BP's assumption that messages are uncorrelated, which results in BP's failure in this regime.

1RSB's approach to incorporating this clustering of solutions into the
equilibrium conditions is to define a new Gibbs measure over
clusters. Let $\y \subset \set{S}$ denote a cluster of solutions and
$\set{Y}$ be the set of all such clusters.  The idea is to treat
$\set{Y}$ the same as we treated $\set{X}$, by defining a distribution
\begin{align}\label{eq:py}
  \psp(\y) \quad \propto \quad {|\y|^\parisi} \quad \forall \; \y \in \set{Y}
\end{align}
where $\parisi \in [0,1]$, called the Parisi parameter
\citep{Mezard1987}, specifies how each cluster's weight depends
on its size.  This implicitly defines a distribution over $\set{X}$
\begin{align}\label{eq:py2px}
  \p(\x) \quad \propto \quad \sum_{ \y \ni \x} \psp(\y)
\end{align}
N.b., $\parisi = 1$ corresponds to the original distribution (Eq~\ref{eq:p}).

\begin{example}
  Going back to our simple 3-SAT example, $\y^{(1)} = \{(True, True,
  True) \}$ and $ \y^{(2)} = \{ (False, False, False) ,\ (False,
  False, True) \}$ are two clusters of solutions.  Using $m = 1$, we
  have \\$\psp(\{\{True, True, True\} \}) = 1/3$ and
  $\psp(\{ \{False, False, False \} , \{ False, False, True\} \}) =
  2/3$.  This distribution over clusters reproduces the distribution
  over solutions -- \ie $\p(\x) = 1/3 \; \forall x \in \set{S}$.  On
  the other hand, using $m = 0$, produces a uniform distribution over
  clusters, but it does not give us a uniform distribution over the
  solutions.
\end{example}

This meta-construction for $\p(\y)$ can be represented using an
auxiliary PGM. One
may use BP to find marginals over this PGM; here BP messages are
distributions over all BP messages in the original PGM, as each
cluster is a fixed-point for BP.  This requirement to represent a
distribution over distributions makes 1RSB practically intractable.
In general, each original BP message is a distribution over
$\set{X}_i$ and it is difficult to define a distribution over this
infinite set.  However this simplifies if the original BP messages can
have limited values.  Fortunately if we apply max-product BP to solve
a CSP, instead of sum-product BP (of Eqs~\ref{eq:miI} and \ref{eq:mIi}), the
messages can have a finite set of values.

\vspace{.5in}

\noindent \textbf{Max-Product BP:} Our previous formulation of CSP was
using sum-product BP.  In general, max-product BP is used to find the
Maximum a Posteriori (MAP) assignment in a PGM, which is a single assignment
with the highest probability.  In our PGM, the MAP assignment is a
solution for the CSP.  The max-product update equations are
\begin{eqnarray}
  &\Msi{i}{{I}}(\x_i) \; = \; \prod_{J \in \partial i \back I}
  \Msi{J}{i}(\x_i) \; &= \; \mbpiI{i}{I}(\ \{\Msi{J}{i}\}_{J \in \partial i \back I}\ )(\x_i)\label{eq:miImp}\\
  &\Msi{I}{i}(\x_i) \; = \; \max_{
    \set{X}_{\partial I \back i}} \f_{I}(\x_I) \prod_{j \in \partial I \back i} \Msi{j}{I}(\x_j) \; &= \;
  \mbpiI{I}{i}(\ \{\Msi{j}{I}\}_{j \in \partial I \back i}\ )(\x_i)
  \label{eq:mIimp}\\
  &\ph(\x_{i}) \; = \; \prod_{J \in \partial i} \Msi{J}{i}(\x_i)\;
  &=\; \mbpi{i}(\ \{\Msi{J}{i}\}_{J \in \partial i}\ )(\x_i)
  \label{eq:bpmargmp}
\end{eqnarray}
where $\mbp = \{ \mbpiI{i}{I}, \mbpiI{I}{i}\}_{i, I \in \partial I}$
is the max-product BP operator and $\mbp_i$ represents the marginal estimate as a function of messages.
% where we use the functions $\mathbf{f}(.)$, $\mathbf{g}(.)$ and
% $\mathbf{h}(.)$ to make explicit the fact that updates and beliefs
% are functions of incoming messages.
Note that here messages and marginals are not distributions. We
initialize $\Ms{i}{I}(\x_i) \in \{0,1\}, \; \forall I, i \in \partial
I, \x_i \in \set{X}_i$.  Because of the way constraints and update
equations are defined, at any point during the updates we have
$\Ms{i}{I}(\x_i) \in \{0,1\}$.  This is also true for
$\ph(\x_i)$. Here any of $\Ms{i}{I}(x_i) = 1$, $\Ms{I}{i}(x_i) = 1$ or
$\ph(\x_i) = 1$, shows that value $x_i$ is allowed according
to a message or marginal, while $0$
forbids that value.  Note that $\ph(\x_i) = {0}\; \forall \x_i
\in \set{X}_i$ \emph{iff} no solution was found, because the incoming
messages were contradictory.
% After convergence, we can use a decimation procedure to find a
% solution as each marginal $\ph_i$ may allow multiple assignments for
% each variable.
The non-trivial fixed-points of max-product BP define quasi-solutions
in 1RSB phase, and therefore define clusters $\y$.
\begin{example}
  If we initialize all messages to $1$ for our simple 3-SAT example,
  the final marginals over all the variables are equal to 1, allowing
  all assignments for all variables.  However beside this trivial
  fixed-point, there are other fixed points that correspond to two
  clusters of solutions.

  For example, considering the cluster $\{ (False, False, False) ,
  (False, False, True) \}$, the following $\{ \Msi{i}{I} \}$ (and
  their corresponding $\{ \Msi{I}{i} \}$ %given by Eq~\ref{eq:mIimp})
  define a fixed-point for max-product BP:
  \begin{align*}
    &\Msi{1}{I}(True) = \ph_1(True) = 0 &\quad \Msi{1}{I}(False) = \ph_1(False) = 1 \quad &\forall I \in \partial 1\\
    &\Msi{2}{I}(True) = \ph_2(True) = 0 &\quad \Msi{2}{I}(False) = \ph_2(False) = 1 \quad &\forall I \in \partial 2\\
    &\Msi{3}{I}(True) = \ph_3(True) = 1 &\quad \quad \Msi{3}{I}(False)
    = \ph_3(False) = 1 \quad& \forall I \in \partial 3
  \end{align*}
  Here the messages indicate the allowed assignments within this
  particular cluster of solutions.
\end{example}

%\vspace{.5in}

\subsubsection{Survey Propagation}\label{sec:sp}
Here we define the 1RSB update equations over max-product BP messages.  We skip
the explicit construction of the auxiliary PGM that results in SP
update equations, and confine this section to the intuition offered by
SP messages. For the construction of the auxiliary-PGM see
\citep{Braunstein2003} and \citep{Mezard2009}.  See~\citep{maneva_new_2004}
for a different perspective on the relation
of BP and SP for the satisfiability problem and \citep{kroc2012survey}
for an experimental study of SP applied to SAT.

Let $\set{Y}_i = 2^{|\set{X}_i|}$ be the power-set\footnote{The
  power-set of $\set{X}$ is the set of all subsets of $\set{X}$,
  including $\{\}$ and $\set{X}$ itself.} of $\set{X}_i$.  Each
max-product BP message can be seen as a subset of $\set{X}_i$ that
contains the allowed states. Therefore $\set{Y}_i$ as its power-set
contains all possible max-product BP messages.  Each message
$\Msp{i}{I}:\; \set{Y}_i \to [0,1]$ in the auxiliary PGM defines a
\emph{distribution} over original max-product BP messages.

\begin{example}[3-COL] $\set{X}_i = \{ 1,2,3 \}$ is the set of colors
  and \\$\set{Y}_i = \{ \{\},\{ 1\},\{ 2\},\{ 3\},\{ 1,2\},\{ 2,3\},\{
  1,3\},\{ 1,2,3\}\}$.  Here
  $\y_i = \{ \}$ corresponds to the case where none of the colors are
  allowed.
\end{example}

Applying sum-product BP to our auxiliary PGM gives entropic
SP($\parisi$) updates as:
\begin{align}
  &\Msp{i}{I}(\y_i) \propto |\y_i|^{\parisi} \;
  \sum_{\{\Msi{J}{i}\}_{J \in \partial i \back I}} \delta(\y_i,
  \mbpiI{i}{I}(\{\Msi{J}{i}\}_{J \in \partial i \back I}))
  \prod_{J \in \partial i \back I} \Msp{J}{i}(\Msi{J}{i}) \label{eq:spiI}\\
  &\Msp{I}{i}(\y_i) \propto |\y_i|^{\parisi} \;
  \sum_{\{\Msi{j}{I}\}_{j \in \partial I \back i}} \delta(\y_i,
  \mbpiI{I}{i}(\{\Msi{j}{I}\}_{j \in \partial I \back i}))
  \prod_{j \in \partial I \back i} \Msp{j}{I}(\Msi{j}{I})  \label{eq:spIi}\\
  &\Msp{i}{{I}}(\{ \}) \quad := \quad \Msp{I}{{i}}(\{ \}) \quad :=
  \quad 0\quad \quad \forall i,\; I \in \partial i \label{eq:nocontribution}
\end{align}
where the summations are over all combinations of max-product BP
messages.  Here the $\delta$-function ensures that only the set of
incoming messages that satisfy the original BP equations make
contributions.  Since we only care about the valid assignments and
$\y_i = \{\}$ forbids all assignments, we ignore its contribution (Eq~\ref{eq:nocontribution}).

\begin{example}[3-SAT]
Consider the SP message $\Msp{1}{C_1}(\y_1)$ in the factor graph of \refFigure{fig:3satfg}.
Here the summation in Eq~\ref{eq:spiI} is over all possible combinations of incoming max-product BP messages
 $\Msi{C_2}{1},\ldots,\Msi{C_5}{1}$. Since each of these messages can assume one of the
three valid values -- \eg \\$\Msi{C^2}{1}(\x_1) \in \{\  \{True\},\{False\},\{True, False\}\  \}$ -- for each particular assignment of $\y_1$,
a total of \\
$|\{ \{True\},\{False\},\{True, False\}\}|^{|\partial 1 \back C_1|} = 3^{4}$
possible combinations are enumerated in the summations of Eq~\ref{eq:spiI}.
However only the combinations that form a valid max-product message update have non-zero
contribution in calculating $\Msp{1}{C_1}(\y_1)$. These are basically the messages that
appear in a max-product fixed point as discussed in Example 8.
\end{example}

Each of original messages corresponds to a different sub-set of clusters
and $\parisi$ (from Eq~\ref{eq:py}) controls the effect of each
cluster's size on its contribution.  At any point, we can use these
messages to estimate the marginals of $\psph(\y)$ defined in
Eq~\ref{eq:py}
\begin{align}
  \psph(\y_i)\; \propto \; |\y_i|^{\parisi} \; \sum_{\{\Msi{J}{i}\}_{J
      \in \partial i}} \delta(\y_i,\ \mbpi{i}(\ \{\Msi{J}{i}\}_{J
    \in \partial i})\ ) \prod_{J \in \partial i}
  \Msp{J}{i}(\Msi{J}{i})\label{eq:spmarg}
\end{align}

This also implies a distribution over the original domain, which we
slightly abuse notation to denote by:
\begin{align}\label{eq:pyi2pxi}
  \psph(\x_i) \; \propto \; \sum_{\y_i \ni \x_i} \psph(\y_i)
\end{align}

 The term SP usually refers to SP($0$) -- \ie $m = 0$ -- where all clusters, regardless of their size,
contribute the same amount to $\psp(\y)$.
Now that we can obtain an estimate of marginals, we can employ this procedure
within a decimation process to incrementally fix some variables.
Here either $\psph(\x_i)$ or $\psph(\y_i)$ can be used by the decimation procedure
to fix the most biased variables. In the former case, a variable $\y_i$ is fixed
to $\y_i^* = \{\x^*_i\}$ when $\x^*_i = \arg_{x_i}\max \psph(\x_i)$.
In the latter case, $\y_i^* = \arg_{y_i} \max \psph(\y_i)$.
Here we use SP-dec(S) to refer to the former procedure (that uses $\psph(\x_i)$
to fix variables to a \emph{single} value) and use SP-dec(C) to refer to the
later case (in which variables are fixed to a \emph{cluster} of assignments).

The original decimation procedure
for $\kappa$-SAT \citep{braunstein_survey_2002} corresponds to  SP-dec(S).  SP-dec(C) for
 CSP with Boolean variables is only slightly different, as SP-dec(C) can choose to fix a cluster to $\y_i = \{ True, False\}$ in addition to the options of
$\y_i = \{True\}$ and $\y_i = \{False\}$, available to SP-dec(S).
However, for larger domains (\eg $q$-COL),
SP-dec(C) has a clear advantage. For example in $3$-COL, SP-dec(C) may choose to fix a cluster to $\y_i = \{1,2\}$ while SP-dec(S) can only choose
between $\y_i \in \{ \{1\},\{2\}, \{3\}\}$. This significant difference is also reflected in their comparative success-rate on $q$-COL.\footnote{Previous applications of SP-dec to $q$-COL by \cite{braunstein2003polynomial} used a heuristic for decimation that is similar SP-dec (C).} (See \refTable{table:rcsp} in \refSection{sec:results}.)

During the decimation process, usually after fixing a subset of variables,
the SP marginals $\psph(\x_i)$ become uniform, indicating that clusters of solutions
have no preference over particular assignments of the remaining variables.
The same happens when we apply SP to random instances in
 RS phase. At this point (\aka paramagnetic phase),
a local search method or BP-dec can often efficiently find an assignment to the
variables that are not yet fixed by decimation.
Note that both SP-dec(C) and SP-dec(S) switch to local search as soon as all $\psph(\x_i)$ become close to uniform.

The computational complexity of each SP update of Eq~\ref{eq:spIi} is
$\set{O}(2^{|\set{X}_i|} - 1)^{|\partial I|}$ as for each particular value $\y_i$, SP needs to consider
every combination of incoming messages, each of which can take
$2^{|\set{X}_i|}$ values (minus the empty set). Similarly, using a naive approach the cost of update of Eq~\ref{eq:spiI} is
$\set{O}(2^{|\set{X}_i|} - 1)^{|\partial i|}$. However by considering incoming messages one at a time, we can perform the same exact update in $\set{O}(|\partial i| \; 2^{2 |\set{X}_i|})$.
In comparison to the cost of
BP updates, we see that SP updates are substantially more expensive
for large $|\set{X}_i|$ and $|\partial I|$.\footnote{Note that our
  representation of distributions is over-complete -- that is we are not
  using the fact that the distributions sum to one. However even in their
  more compact forms, for general CSPs, the cost of each SP update
  remains exponentially larger than that of BP (in $|\set{X}_i|$,
  $|\partial I|$). However if the factors are sparse and have high order,
both BP and SP allow more efficient updates.}

\subsection{Perturbed Survey Propagation}\label{sec:psp}
The perturbation scheme that we use for SP is similar to what we did
for BP. Let \\$\sppiI{i}{I}(\ \{ \Msp{j}{J}\}_{j \in \mb i, (J \in \partial i) \back
   I})\ )$ denote the update operator for the message from variable $\y_i$
 to factor $\f_I$. This operator is obtained by substituting
Eq~\ref{eq:spIi} into Eq~\ref{eq:spiI} to get a single SP update
equation.
Let $\spp (\{\Msp{i}{I}\}_{i, I \in \partial i})$
% \triangleq \{
% \sppiI{i}{I}( \{ \Msp{j}{J}\}_{j \in \mb i, (J \in \partial i) \back
%   I}) ) \}_{i, I \in \partial i}$
denote the aggregate SP operator, which applies $\sppiI{i}{I}$ to update each individual
message.

We perform Gibbs sampling from the ``original'' domain $\set{X}$ using the
implicit marginal of Eq~\ref{eq:pyi2pxi}.  We denote this random
operator by $\gssp = \{ \gsspi{i}\}_i$:
\begin{align}
  \Msp{i}{I}(\y_i) \;=\; \gsspi{i}(\ \{\Msp{j}{J}\}_{j \in \mb i, J \in \partial i}\ ) \; \triangleq \;
  \delta(\y_i, \{\xh_i\}) \quad \text{where} \; \xh_i \sim
  \psph(\x_i)
\end{align}
where the second argument of the $\delta$-function is a singleton set, containing
a sample from the estimate of marginal.

Now, define the Perturbed SP operator as the convex combination of SP
and either of the GS operator above:
\begin{align}
%  \pspp^C(\{\Msp{i}{I}\}) \triangleq \gamma \gssp^C(\{\Msp{i}{I}\}) + (1 - \gamma) \spp(\{\Msp{i}{I}\}) \quad \text{Perturbed SP(C)}\\
  \pspp(\{\Msp{i}{I}\}) \; \triangleq \; \gamma \gssp(\{\Msp{i}{I}\})\; + \; (1 - \gamma) \spp(\{\Msp{i}{I}\}) %\quad
%\text{Perturbed SP(S)}
\end{align}

Similar to perturbed BP, during iterations of Perturbed SP, $\gamma$ is gradually increased from
$0$ to $1$.
%The
%difference between two variations of Perturbed SP is that $\pspp^C$ (Perturbed SP(C))
%produces a cluster of assignments, while $\pspp^S$ (Perturbed SP(S))
%produces a single assignment.
If perturbed SP reaches the final iteration, the samples from the implicit marginals represent a satisfying assignment.
The advantage of this scheme to SP-dec is that perturbed SP does not require any further local search.
In fact we may apply
$\pspp$ to CSP instances in the RS phase as well, where the
solutions form a single giant cluster.  In contrast, applying
SP-dec, to these instances simply invokes the local search method.

To demonstrate this, we applied Perturbed SP(S) to benchmark CSP instances of \refTable{table:benchmark} in which the maximum number of elements in the factor was less than $10$.
Here Perturbed SP(S) solved $80$ instances out of $202$ cases, while Perturbed BP solved $78$ instances.

\subsection{Experiments on random CSP}\label{sec:results}
We implemented all the methods above for general factored CSP using the libdai code base
\citep{libdai}. To our knowledge this is the first general
implementation of SP and SP-dec. Previous applications of SP-dec to $\kappa$-SAT and
$q$-COL \citep{braunstein2003polynomial,mulet_coloring_2002,braunstein_survey_2002}
were specifically tailored to just one of those problems.

\begin{figure}
  \centering
\hbox{
    \includegraphics[width=.507\textwidth]{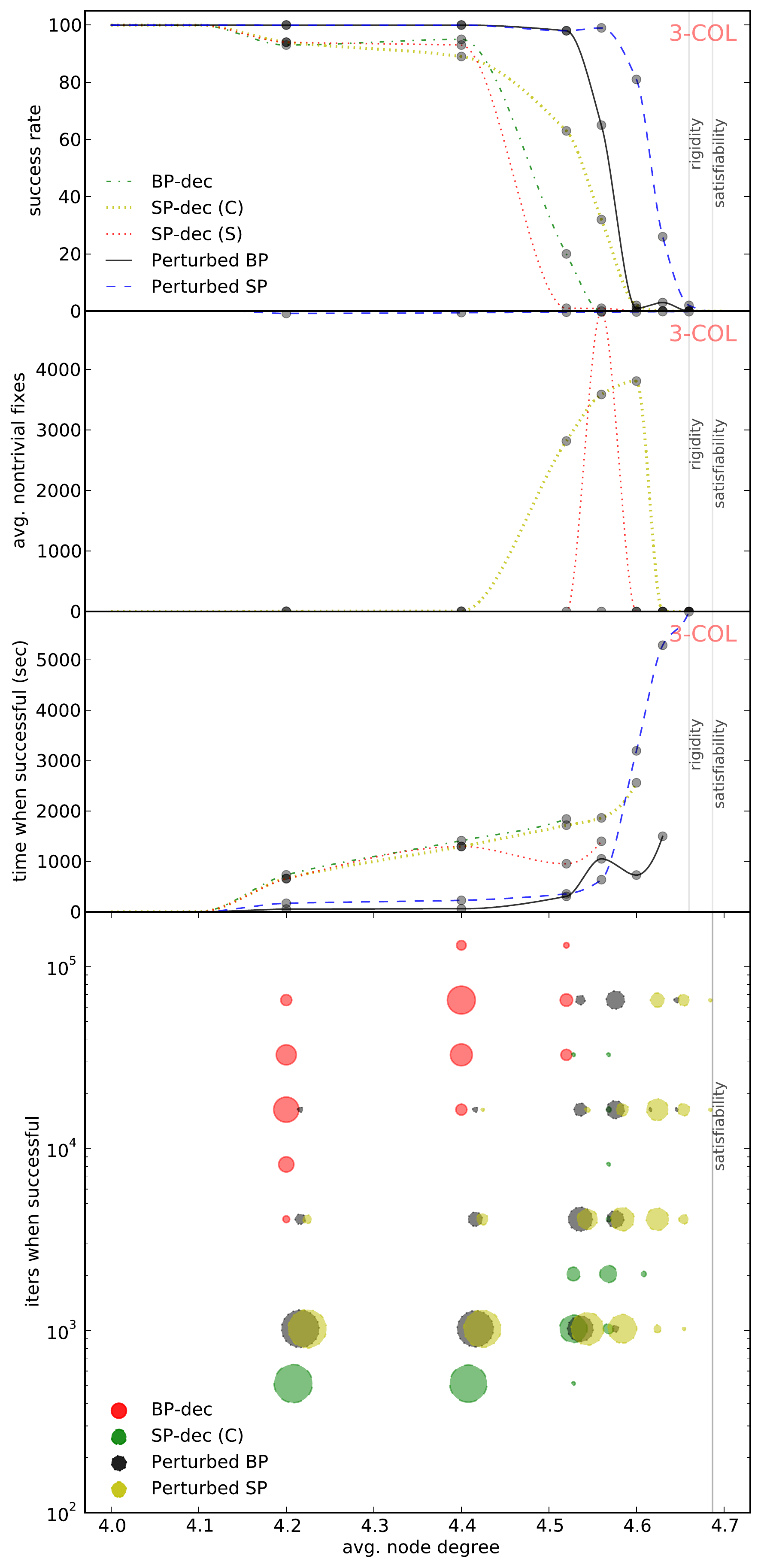}
    \includegraphics[width=.5\textwidth]{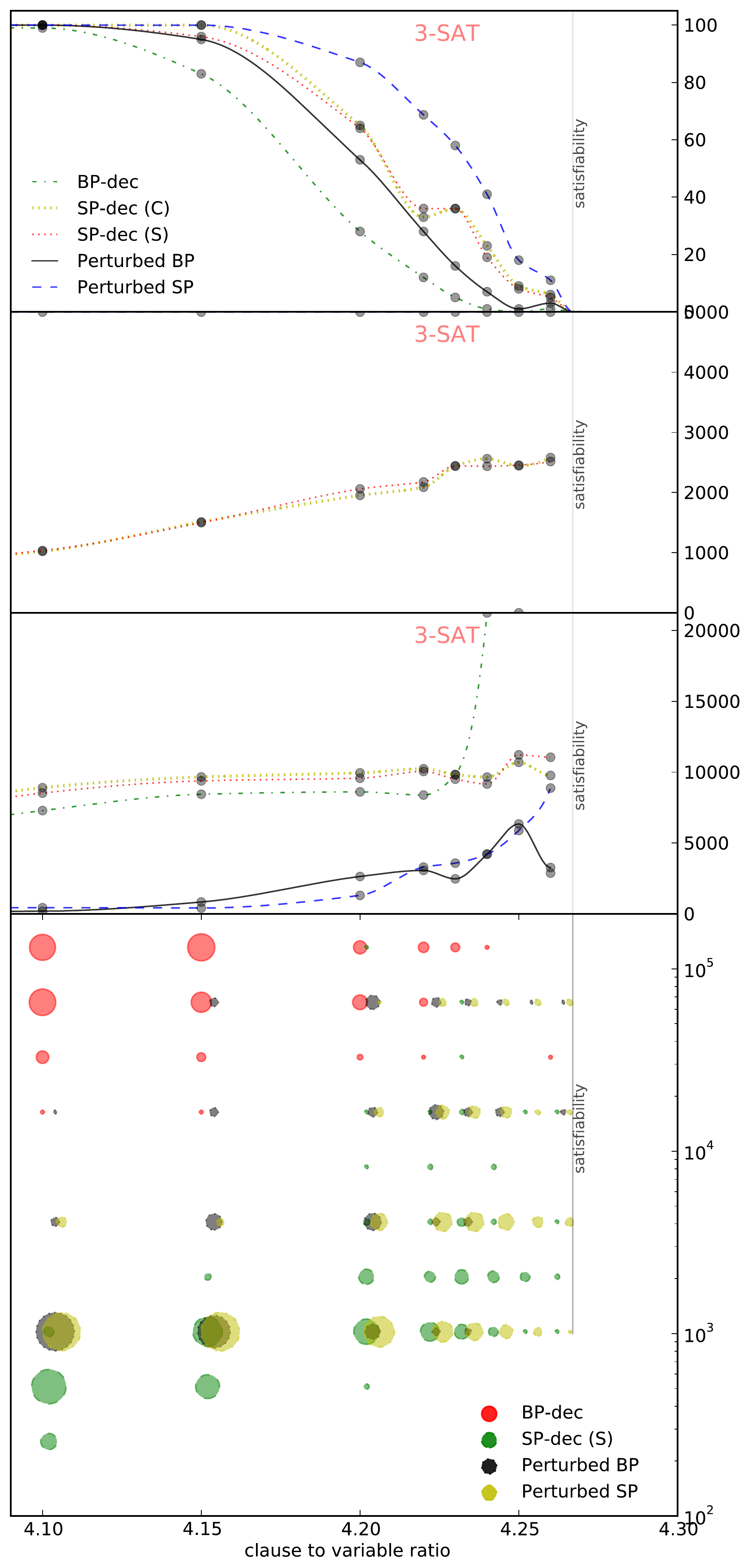}
}
\caption{\small{\textbf{(first row)}~Success-rate of different methods for 3-COL and 3-SAT for various control parameters.
\textbf{(second row)}~The average number of variables (out of $N = 5000$) that are fixed using SP-dec (C) and (S) before calling local search, averaged over 100 instances.
\textbf{(third row)}~The average amount of time (in seconds) used by the successful setting of each method to find a satisfying solution. For SP-dec(C) and (S) this includes the time used by local search.
\textbf{(fourth row)}~The number of iterations used by different methods at different control parameters, when the method
was successful at finding a solution. The number of iterations
for each of 100 random instances is rounded to the closest power of 2. This does not include the iterations used by local search
after SP-dec.
}}
  \label{fig:all3sat3col}
\end{figure}

Here we report the results on $\kappa$-SAT for $\kappa \in \{3,4\}$
and $q$-COL for $q \in \{3,4,9\}$.  We used the procedure discussed
in the examples of \refSection{sec:clustering} to produce $100$ random
instances with $N = 5,000$ variables for each control parameter
$\alpha$. We report the probability of finding a satisfying assignment
for different methods (\ie the portion of $100$ instances that were satisfied by each method).
For coloring instances, to help
decimation, we break the initial symmetry
of the problem by fixing a single variable to an arbitrary value.

For BP-dec and SP-dec, we use a convergence threshold of $\epsilon= .001$ and
fix $\rho = 1\%$ of variables per iteration of decimation.
Perturbed BP and Perturbed SP use $T = 1000$ iterations. Decimation-based methods use a maximum of $T = 1000$ iterations per iteration of decimation.
If any of the methods failed to find a solution in the first attempt, $T$ was increased by a factor of $4$ at most $3$ times (so in the final
attempt: $T = 64,000$). To avoid blow-up in run-time, for BP-dec and SP-dec, only the maximum iteration, $T$, during the first iteration of decimation, was increased
(this is similar to the setting of \citet{braunstein_survey_2002} for SP-dec).
For both variations of SP-dec (see \refSection{sec:sp}), after each decimation step, if $\max_{i,x_i} \p(x_i) - \frac{1}{|\set{X}_i|} < .01$
we consider the instance para-magnetic,
and run BP-dec (with $T = 1000$, $\epsilon = .001$ and $\rho = 1\%$) on the simplified instance.
%For Perturbed SP(C) (which recall finds a cluster of assignments), we use Perturbed BP with $T = 1000$ as the local search method.

\refFigure{fig:all3sat3col}(first row) visualizes the success rate of different methods on $100$ instances of 3-SAT (right) and 3-COL (left).
\refFigure{fig:all3sat3col}(second row) reports the number of variables that are fixed by SP-dec(C) and (S) before calling BP-dec as local search.
The third row shows the average amount of time that is used to find a satisfying solution. This does not include the failed attempts.
For SP-dec variations, this time includes the time used by local search.
The final row of \refFigure{fig:all3sat3col} shows the number of iterations used by each
method at each level of difficulty over the successful instances.
Here the area of each disk is proportional
to the frequency of satisfied instances with that particular number of iterations for each control parameter and inference method\footnote
{The number of iterations are rounded to the closest power of two.}.

Here we make the following observations:
\begin{itemize}
\item \textbf{Perturbed BP is much more effective than BP-dec, while remaining ten to hundreds of times
 more efficient.}
\item As the control parameter grows larger,
the chance of requiring more iterations to satisfy the instance increases for all methods.
\item Although computationally very inefficient, BP-dec is able to find solutions for instances
with larger control parameters than suggested by previous results \citep[\eg][]{Mezard2009}.
\item For many instances where SP-dec(C) and (S) use few iterations,
the variables are fixed to a trivial cluster
$\y_i = \set{X}_i$,  which allows all assignments.
This is particularly pronounced for 3-COL, where up to $\alpha = 4.4$ the non-trivial fixes remains zero
and therefore the success rate up to this point is solely due to  BP-dec.
%Where non-trivial fixes are zero, the success rate is solely due to local search (\ie BP-dec). For 3-COL,
\item While for 3-SAT, SP-dec(C) and SP-dec(S) have a similar performance, for 3-COL,  SP-dec(C) significantly
outperforms SP-dec(S).
\end{itemize}

\refTable{table:rcsp} reports the
success-rate as well as the average of total iterations in the
\emph{successful} attempts of each method. Here the number of iterations
for SP-dec(C) and (S) is the sum of iterations
used by the method and the following local search.
We observe that Perturbed BP can solve most of the easier instances using only $T=1000$ iterations (\eg see Perturb BP's result for 3-SAT at $\alpha = 4.1$, 3-COL at $\alpha = 4.2$ and 9-COL at $\alpha = 33.4$).

\refTable{table:rcsp} also supports our speculation in \refSection{sec:sp} that \textbf{SP-dec(C) is in general preferable to SP-dec(S)}, in particular when applied to the coloring problem.

The most important advantage of Perturbed BP over SP-dec and Perturbed SP is that Perturbed BP can be applied to instances with large factor cardinality (\eg $10$-SAT) and
large variable domains (\eg $9$-COL). For example for $9$-COL, the cardinality of each SP message is $2^9=512$, which makes SP-dec and Perturbed SP impractical.
Here BP-dec is not even able to solve a single instance around the dynamical transition (as low as $\alpha = 33.4$) while Perturbed BP satisfies all instances up to $\alpha = 34.1$.\footnote{
Note that for $9$-COL condensation transition happens after rigidity transition. So if we were able to find solutions after
rigidity, it would have implied that condensation transition marks the onset of difficulty. However, this did not occur and similar to all other cases, Perturbed BP failed
before rigidity transition.}
Besides the experimental results reported here,
we have also used perturbed BP to efficiently solve other CSPs such as K-Packing, K-set-cover and
clique-cover within the context of min-max inference~\citep{ravanbakhsh_minmax}.

\begin{table}[!pht]
%\begin{table}
%\begin{adjustbox}{angle=90}
  \caption[Comparison of different methods for K-satisfiability and K-coloring.]{Comparison of different methods on  \textbf{$\{3,4\}$-SAT} and \textbf{$\{3,4,9\}$-COL}.
For each method the success-rate and the average  number of iterations (including local search) on successful attempts are reported.
The approximate location of phase transitions are from \citep{montanari_clusters_2008,zdeborova_phase_2007-4}
.}\label{table:rcsp}
\centering
\begin{tikzpicture}
\node (table) [inner sep=0pt] {
    \scalebox{.6}{
    \begin{tabu}{c  c |[2pt] r  l |[2pt] r  l |[2pt] r  l|[2pt] r  l|[2pt] r  l}
      \cline{3-12}
      & &\multicolumn{2}{ c |[2pt]}{BP-dec}& \multicolumn{2}{ c|[2pt] }{SP-dec(C)}&\multicolumn{2}{ c|[2pt] }{SP-dec(S)}&
      \multicolumn{2}{ c|[2pt] }{Perturbed BP}& \multicolumn{2}{ c }{Perturbed SP}\\
      \cline{3-12}
      \begin{sideways}Problem \end{sideways}&
      \begin{sideways}ctrl param $\alpha$\end{sideways}&
      \begin{sideways}avg. iters.\end{sideways} &
      \begin{sideways}success rate\end{sideways} &
      \begin{sideways}avg. iters.\end{sideways} &
      \begin{sideways}success rate\end{sideways} &
      \begin{sideways}avg. iters.\end{sideways} &
      \begin{sideways}success rate\end{sideways} &
      \begin{sideways}avg. iters.\end{sideways} &
      \begin{sideways}success rate\end{sideways} &
      \begin{sideways}avg. iters.\end{sideways} &
      \begin{sideways}success rate\end{sideways}
      \\\tabucline[2pt]{-}
      \multirow{10}{*}{\textbf{3-SAT}}&
      3.86 & \multicolumn{10}{l}{dynamical and condensation transition}\\\cline{2-12}
&4.1 & 85405 & 99\% &102800 & 100\% &96475 & 100\% &1301 & 100\%  &1211 & 100\%
%4.1 & 81196 & 100\% &96055 & 100\% &95113 & 100\% &1179 & 100\% &2061 & 100\% &1060 & 100\%
\\\cline{2-12}
&4.15 & 104147 & 83\% &118852 & 100\% &111754 & 96\% &5643 & 95\%  &1121 & 100\%
%4.15 & 98808 & 77\% &106367 & 98\% &105675 & 98\% &2801 & 95\% &4351 & 83\% &1215 & 98\%
\\\cline{2-12}
&4.2 & 93904 & 28\% &118288 & 65\% &113910 & 64\% &19227 & 53\%  &3415 & 87\%
%4.2 & 95691 & 31\% &110975 & 79\% &96730 & 82\% &18460 & 61\% &9624 & 37\% &4799 & 94\%
\\\cline{2-12}
&4.22 & 100609 & 12\% &112910 & 33\% &114303 & 36\% &22430 & 28\%  &8413 & 69\%
%4.22 & 145060 & 3\% &111239 & 44\% &101184 & 41\% &32351 & 20\% &23002 & 4\% &10640 & 60\%
\\\cline{2-12}
&4.23 & 123318 & 5\% &109659 & 36\% &107783 & 36\% &18438 & 16\%  &9173 & 58\%
%4.23 & 126089 & 2\% &98843 & 34\% &91032 & 36\% &36751 & 12\% &18502 & 4\% &10301 & 50\%
\\\cline{2-12}
&4.24 & 165710 & 1\% &126794 & 23\% &118284 & 19\% &29715 & 7\%  &10147 & 41\%
%4.24 & 59283 & 1\% &105008 & 21\% &96602 & 20\% &33144 & 7\% &26002 & 4\% &9369 & 38\%
\\\cline{2-12}
&4.25 & N/A & 0\% &123703 & 9\% &110584 & 8\% &64001 & 1\%  &14501 & 18\%
%4.25 & N/A & 0\% &66509 & 11\% &92478 & 13\% &48251 & 4\% &65002 & 1\% &16626 & 24\%
\\\cline{2-12}
&4.26 & 37396 & 1\% &83231 & 6\% &106363 & 5\% &32001 & 3\%  &22274 & 11\%
%4.26 & 37396 & 1\% &83231 & 6\% &106363 & 5\% &32001 & 3\% &2002 & 1\% &22274 & 11\%
\\\cline{2-12}
%-------------------
      &4.268 & \multicolumn{10}{l}{satisfiability transition}
      \\\tabucline[2pt]{-}
      \multirow{7}{*}{\textbf{4-SAT}}&
      9.38 & \multicolumn{10}{l}{dynamical transition}\\\cline{2-12}
      &9.547& \multicolumn{10}{l}{condensation transition}\\\cline{2-12}
&9.73 & 134368 & 8\% &119483 & 32\% &120353 & 35\% &25001 & 43\%  &11142 & 86\% \\\cline{2-12}
&9.75 & 168633 & 5\% &115506 & 15\% &96391 & 21\% &36668 & 27\%  &9783 & 68\% \\\cline{2-12}
&9.78 & N/A & 0\% &83720 & 9\% &139412 & 7\% &34001 & 12\%  &11876 & 37\% \\\cline{2-12}
      &9.88& \multicolumn{10}{l}{ rigidity transition}\\\cline{2-12}
      &9.931& \multicolumn{10}{l}{ satisfiability transition}
      \\\tabucline[2pt]{-}
      \multirow{10}{*}{\textbf{3-COL}}&
      4 & \multicolumn{10}{l}{dynamical and condensation transition}\\\cline{2-12}
&4.2 & 24148 & 93\% &25066 & 94\% &24634 & 94\% &1511 & 100\%  &1151 & 100\%
%4.2 & 24598 & 94\% &21675 & 98\% &20414 & 92\% &1391 & 100\% &2542 & 100\% &1181 & 100\%
\\\cline{2-12}
&4.4 & 51590 & 95\% &52684 & 89\% &54587 & 93\% &1691 & 100\%  &1421 & 100\%
%4.4 & 46996 & 96\% &47829 & 93\% &45314 & 96\% &2728 & 99\% &2662 & 100\% &1421 & 100\%
\\\cline{2-12}
&4.52 & 61109 & 20\% &68189 & 63\% &54736 & 1\% &7705 & 98\%  &2134 & 98\%
%4.52 & 22252 & 13\% &59469 & 70\% &N/A & 0\% &8066 & 93\% &8264 & 80\% &3191 & 100\%
\\\cline{2-12}
&4.56 & N/A & 0\% &63980 & 32\% &13317 & 1\% &28047 & 65\%  &3607 & 99\%
%4.56 & 11978 & 6\% &56191 & 40\% &N/A & 0\% &26598 & 62\% &10079 & 39\% &4062 & 98\%
\\\cline{2-12}
&4.6 & N/A & 0\% &74550 & 2\% &N/A & 0\% &16001 & 1\%  &18075 & 81\%
%4.6 & N/A & 0\% &22394 & 4\% &N/A & 0\% &50501 & 8\% &20302 & 20\% &20037 & 84\%
\\\cline{2-12}
&4.63 & N/A & 0\% &N/A & 0\% &N/A & 0\% &48001 & 3\%  &29270 & 26\%
%4.63 & N/A & 0\% &N/A & 0\% &N/A & 0\% &N/A & 0\% &39502 & 10\% &34241 & 25\%
\\\cline{2-12}
     &4.66& \multicolumn{10}{l}{rigidity transition}\\\cline{2-12}
&4.66 & N/A & 0\% &N/A & 0\% &N/A & 0\% &N/A & 0\%  &40001 & 2\%
%4.66 & N/A & 0\% &N/A & 0\% &N/A & 0\% &N/A & 0\% &N/A & 0\% &16001 & 2\%
\\\cline{2-12}
      &4.687& \multicolumn{10}{l}{ satisfiability transition}
      \\\tabucline[2pt]{-}
      \multirow{6}{*}{\textbf{4-COL}}&
      8.353 & \multicolumn{10}{l}{dynamical  transition}\\\cline{2-12}
&8.4 & 64207 & 92\% &72359 & 88\% &71214 & 93\% &1931 & 100\% &1331 & 100\% \\\cline{2-12}
&8.46 & \multicolumn{10}{l}{dynamical  transition}\\\cline{2-12}
&8.55 & 77618 & 13\% &60802 & 13\% &62876 & 9\% &3041 & 100\%  &5577 & 100\% \\\cline{2-12}
&8.7 & N/A & 0\% &N/A & 0\% &N/A & 0\% &50287 & 14\% &N/A  & 0\% \\\cline{2-12}
      &8.83& \multicolumn{10}{l}{rigidity transition}\\\cline{2-12}
      &8.901& \multicolumn{10}{l}{ satisfiability transition}
      \\\tabucline[2pt]{-}
      \multirow{9}{*}{\textbf{9-COL}}&
33.45 & \multicolumn{10}{l}{dynamical  transition}\\\cline{2-12}
&33.4 & N/A & 0\%& N/A&N/A&N/A&N/A &1061 & 100\%&N/A&N/A \\\cline{2-12}
&33.9 & N/A & 0\%& N/A&N/A&N/A&N/A &3701 & 100\%&N/A&N/A \\\cline{2-12}
&34.1 & N/A & 0\%& N/A&N/A&N/A&N/A &12243 & 100\%&N/A&N/A \\\cline{2-12}
&34.5 & N/A & 0\%& N/A&N/A&N/A&N/A &48001 & 6\%&N/A&N/A \\\cline{2-12}
&35.0 & N/A & 0\%& N/A&N/A&N/A&N/A & N/A & 0\%&N/A&N/A \\\cline{2-12}
      &39.87& \multicolumn{10}{l}{rigidity transition}\\\cline{2-12}
      &43.08& \multicolumn{10}{l}{condensation transition}\\\cline{2-12}
      &43.37& \multicolumn{10}{l}{ satisfiability transition}
    \end{tabu}
}
};
\draw [rounded corners=.5em] (table.north west) rectangle (table.south east);
\end{tikzpicture}
%\end{adjustbox}
\end{table}

\subsection{Discussion}\label{sec:discussion}

It is easy to check that, for $\parisi = 1$, SP updates produce
sum-product BP messages as an average case; that is, the SP updates
(Eqs~\ref{eq:spiI} and \ref{eq:spIi}) reduce to that of sum-product BP
(Eqs~\ref{eq:miI} and \ref{eq:mIi}) where
\begin{align}
  \Ms{i}{{I}}(\x_i) \quad \propto \sum_{\y_i \ni \x_i}
  \Msp{i}{{I}}(\y_i)
\end{align}

\begin{figure*}
\centering
    \includegraphics[width=.4\textwidth]{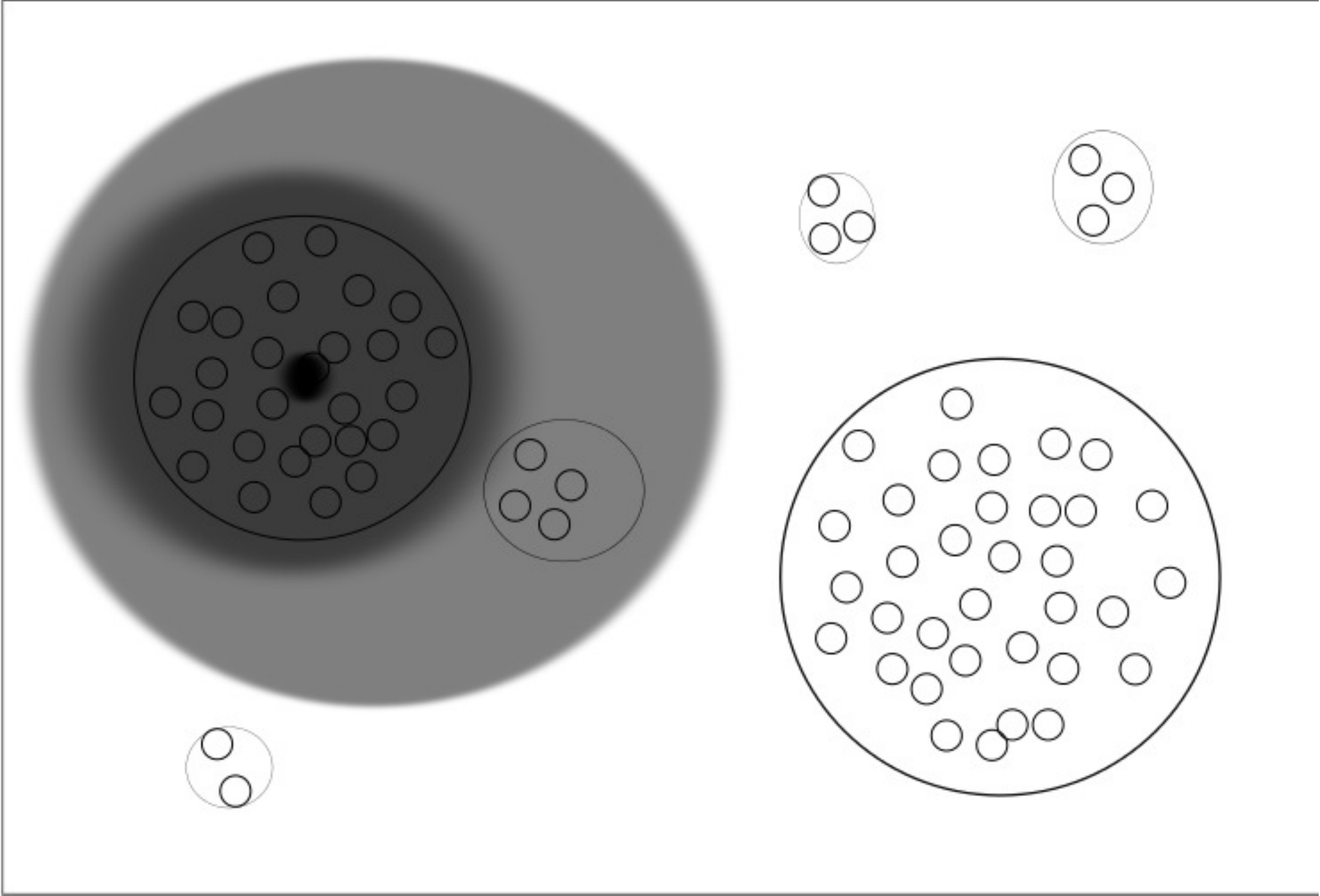}
  \caption{\small{ This schematic view demonstrates the clustering
      during condensation phase.  Here assume $x$ and $y$ axes
      correspond to $\x_1$ and $\x_2$.  Considering the whole space of
      assignments, $\x_1$ and $\x_2$ are highly correlated. The formation of this correlation between distant variables on a PGM breaks BP. Now
      assume that Perturbed BP messages are focused on the largest
      shaded ellipse.  In this case the correlation is significantly
      reduced.  } }\label{fig:1sRSB}
\end{figure*}

This suggests that the BP equation remains correct wherever SP($1$)
holds, which has lead to the belief that BP-dec should perform well up
to the condensation transition \citep{krzakala_gibbs_2007}.
However in reaching this conclusion, the effect of decimation was
ignored.  More recent analyses
\citep{coja-oghlan_belief_2010,montanari_solving_2007,ricci-tersenghi_cavity_2009}
draw a similar conclusion about the effect of decimation: At some
point during the decimation process, variables form long-range
correlations such that fixing one variable may imply an assignment for
a portion of variables that form a loop, potentially leading to
contradictions. Alternatively the same long-range correlations result
in BP's lack of convergence and error in marginals that may lead to
unsatisfying assignments.

Perturbed BP avoids the pitfalls of BP-dec in two ways: \textbf{(I)} Since many
configurations have non-zero probability until the final iteration,
Perturbed BP can avoid contradictions by adapting to the most recent
choices.  This is in contrast to decimation in which variables are
fixed once and are unable to change afterwards.  A backtracking scheme
suggested by \citet{parisi_backtracking_2033} attempts to fix the
same problem with SP-dec.  \textbf{(II)} We speculate that simultaneous bias of
all messages towards sub-regions over which the BP equations remain
valid, prevents the formation of long-range correlations between
variables that breaks BP in 1sRSB; see \refFigure{fig:1sRSB}.

In all experiments, we observed that Perturbed BP is competitive with
SP-dec, while BP-dec often fails on much easier problems.  We saw that
the cost of each SP update
grows exponentially faster than the cost of each BP update.  Meanwhile, our
perturbation scheme adds a negligible cost to that of BP -- \ie that of
sampling from these local marginals and updating the outgoing messages
accordingly.
Considering the computational complexity of
SP-dec, and also the limited setting under which it is applicable,
Perturbed BP is an attractive substitute.
Furthermore our experimental results also suggest that Perturbed SP(S) is a viable
option for real-world CSPs with small variable domains and constraint cardinalities.

\section*{Conclusion}\label{sec:conclusion}
We considered the challenge of efficiently producing assignments that
satisfy hard combinatorial problems, such as $\kappa$-SAT and $q$-COL.
We focused on ways to use message passing methods to solve CSPs, and
introduced a novel approach, Perturbed BP, that combines BP and GS in
order to sample from the set of solutions.  We demonstrated that
Perturbed BP is significantly more efficient and successful than BP-dec.
We also demonstrated that Perturbed BP can be as powerful as
a state-of-the-art algorithm (SP-dec), in solving rCSPs while remaining tractable for problems with large
variable domains and factor cardinalities.  Furthermore
we provided a method to apply the similar perturbation procedure to
SP, producing the Perturbed SP process that outperforms SP-dec in
solving difficult rCSPs.

\section*{Acknowledgment}
We would like to thank anonymous reviewers for their constructive and insightful feedback.
RG received support from NSERC and Alberta Innovate Center for Machine Learning (AICML).
Research of SR has been supported by Alberta Innovates Technology Futures, AICML and QEII graduate scholarship.
This research has been enabled by the use of computing resources provided by WestGrid and Compute/Calcul Canada.

\bibliography{jmlr2}

\appendix
\section*{A. Detailed Results for Benchmark CSP}
In \refTable{table:benchmark} we report the average
number of iterations and average time for the attempt that
successfully found a satisfying assignment (\ie the failed instances
are not included in the average).  We also report the number of
satisfied instances for each method as well as the number of
satisfiable instance in that series of problems (if known).  Further
information about each data-set maybe obtained from \cite{xcsp}.

\begin{table}
  \caption{\small{Comparison of Perturbed BP and BP-guided decimation on benchmark CSPs.}}\label{table:benchmark}
  \begin{center}
    \scalebox{0.7}{
      \begin{tabu}{c c c c |[2pt] c | c| c|[2pt] c| c |l|}
        \cline{5-10}
        & & &&\multicolumn{3}{ c|[2pt] }{BP-dec}& \multicolumn{3}{ c| }{Perturbed BP}\\\hline
        \multicolumn{1}{ c| }{\begin{sideways}problem\end{sideways}} &
        \multicolumn{1}{ |c| }{\begin{sideways}series\end{sideways}} &
        \multicolumn{1}{ c| }{\begin{sideways}instances\end{sideways}} &
        \begin{sideways}$\#$ satisfiable\end{sideways} &
        \begin{sideways}$\#$ satisfied\end{sideways} &
        \begin{sideways}avg. time (s) \end{sideways} &
        \begin{sideways}avg. iters\end{sideways} &
        % \begin{sideways}avg. $\rho$\end{sideways} &
        \begin{sideways}$\#$ satisfied\end{sideways} &
        \begin{sideways}avg. time (s)\end{sideways} &
        \begin{sideways}avg. iters\end{sideways}
        \\\tabucline[2pt]{-}
        \multicolumn{1}{c|}{\multirow{1}{*}{Geometric}} &
        \multicolumn{1}{ |c| }{-} &
        % \multicolumn{1}{ c| }{inst} & sat &bpdec sat&bpdec t&bpdec
        % it &pbpsat&pbp t& pbpiter
        \multicolumn{1}{ c| }{100} & 92 & 77 & 208.63 & 30383 &  81 & .70 & 74\\
        \tabucline[1.2pt]{-}
        \multicolumn{1}{c|}{\multirow{6}{*}{Dimacs}} &
        \multicolumn{1}{ |c| }{aim-50} &
        % \multicolumn{1}{ c| }{inst} & sat &bpdec sat&bpdec t&bpdec
        % it &pbpsat&pbp t& pbpiter
        \multicolumn{1}{ c| }{24} & 16 & 9 & 11.41 & 25344 &  14 & .07 & 181\\
        \cline{2-10}
        &\multicolumn{1}{ |c| }{aim-100} &
        \multicolumn{1}{ c| }{24} & 16 & 8 & 18.2 & 16755 &  11 & .15 & 213\\
        \cline{2-10}
        &\multicolumn{1}{ |c| }{aim-200} &
        \multicolumn{1}{ c| }{24} & N/A & 7 & 401.90 & 160884 &  6 & .17 & 46\\
        \cline{2-10}
        &\multicolumn{1}{ |c| }{ssa} &
        \multicolumn{1}{ c| }{8} & N/A & 4 & .60 & 373.25 &  4 & .50 & 86\\
        \cline{2-10}
        &\multicolumn{1}{ |c| }{jhnSat} &
        \multicolumn{1}{ c| }{16} & 16 & 16 & 5839.86 & 141852 &  13 & 9.82 & 117\\
        \cline{2-10}
        &\multicolumn{1}{ |c| }{varDimacs} &
        \multicolumn{1}{ c| }{9} & N/A & 4 & 2.95 & 715 &  4 & .12 & 18\\
        \tabucline[1.2pt]{-}
        \multicolumn{1}{c|}{\multirow{3}{*}{QCP}} &
        \multicolumn{1}{ |c| }{QCP-10} &
        % \multicolumn{1}{ c| }{inst} & sat &bpdec sat&bpdec t&bpdec
        % it &pbpsat&pbp t& pbpiter
        \multicolumn{1}{ c| }{15} & 10 & 10 & 43.87 & 30054 &  10 & .22 & 51\\
        \cline{2-10}
        &\multicolumn{1}{ |c| }{QCP-15} &
        \multicolumn{1}{ c| }{15} & 10 & 3 & 5659.70 & 600741 & 4 & 9.59 & 530\\
        \cline{2-10}
        &\multicolumn{1}{ |c| }{QCP-25} &
        \multicolumn{1}{ c| }{15} & 10 & 0 & 0 & 0 &  0 & 0 & 0\\
        \tabucline[1.2pt]{-}
        \multicolumn{1}{c|}{\multirow{16}{*}{Graph-Coloring}} &
        \multicolumn{1}{ |c| }{ColoringExt} &
        % \multicolumn{1}{ c| }{inst} & sat &bpdec sat&bpdec t&bpdec
        % it &pbpsat&pbp t& pbpiter
        \multicolumn{1}{ c| }{17} & N/A & 4 & .05 & 103 &  5 & .04 & 25\\
        \cline{2-10}
        &\multicolumn{1}{ |c| }{school} &
        \multicolumn{1}{ c| }{8} & N/A & 0 & N/A & N/A & 5 & 62.86 & 153\\
        \cline{2-10}
        &\multicolumn{1}{ |c| }{myciel} &
        \multicolumn{1}{ c| }{16} & N/A & 5 & .21 & 59 &  5 & .05 & 11\\
        \cline{2-10}
        &\multicolumn{1}{ |c| }{hos} &
        \multicolumn{1}{ c| }{13} & N/A & 5 & 27.34 & 606 &  5 & 10.04 & 37\\
        \cline{2-10}
        &\multicolumn{1}{ |c| }{mug} &
        \multicolumn{1}{ c| }{8} & N/A & 4 & .068 & 313 &  4 & .004 & 11\\
        \cline{2-10}
        &\multicolumn{1}{ |c| }{register-fpsol} &
        \multicolumn{1}{ c| }{25} & N/A & 0 & N/A & N/A &  0 & N/A & N/A\\
        \cline{2-10}
        &\multicolumn{1}{ |c| }{register-inithx} &
        \multicolumn{1}{ c| }{25} & N/A & 0 & N/A & N/A &  0 & N/A & N/A\\
        \cline{2-10}
        &\multicolumn{1}{ |c| }{register-zeroin} &
        \multicolumn{1}{ c| }{14} & N/A & 3 & 5906.16 & 26544 &  0 & N/A & N/A\\
        \cline{2-10}
        &\multicolumn{1}{ |c| }{register-mulsol} &
        \multicolumn{1}{ c| }{49} & N/A & 5 & 59.27 & 418 &  0 & N/A & N/A\\
        \cline{2-10}
        &\multicolumn{1}{ |c| }{sgb-queen} &
        \multicolumn{1}{ c| }{50} & N/A & 7 & 35.66 & 916 & 11 &7.56 & 81\\
        \cline{2-10}
        &\multicolumn{1}{ |c| }{sgb-games} &
        \multicolumn{1}{ c| }{4} & N/A & 1 & .91 & 434 &  1 & .07 & 21\\
        \cline{2-10}
        &\multicolumn{1}{ |c| }{sgb-miles} &
        \multicolumn{1}{ c| }{34} & N/A & 4 & 20.86 & 371 &  2 & 4.20 & 181\\
        \cline{2-10}
        &\multicolumn{1}{ |c| }{sgb-book} &
        \multicolumn{1}{ c| }{26} & N/A & 5 & 1.72 & 444 &  5 & .18 & 39\\
        \cline{2-10}
        &\multicolumn{1}{ |c| }{leighton-5} &
        \multicolumn{1}{ c| }{8} & N/A & 0 & N/A & N/A &  0 & N/A & N/A\\
        \cline{2-10}
        &\multicolumn{1}{ |c| }{leighton-15} &
        \multicolumn{1}{ c| }{28} & N/A & 0 & N/A & N/A &  1 & 106.46 & 641\\
        \cline{2-10}
        &\multicolumn{1}{ |c| }{leighton-25} &
        \multicolumn{1}{ c| }{29} & N/A & 2 & 304.49 & 1516 &  2 & 94.11 & 241\\
        \tabucline[1.2pt]{-}
        \multicolumn{1}{c|}{\multirow{1}{*}{All Interval Series}} &
        \multicolumn{1}{ |c| }{series} &
        % \multicolumn{1}{ c| }{inst} & sat &bpdec sat&bpdec t&bpdec
        % it &pbpsat&pbp t& pbpiter
        \multicolumn{1}{ c| }{12} & 12 & 2 & 4.78 & 11319 &  7 & 1.85 & 520\\
        \tabucline[1.2pt]{-}
        \multicolumn{1}{c|}{\multirow{3}{*}{Job Shop}} &
        \multicolumn{1}{ |c| }{e0ddr1} &
        % \multicolumn{1}{ c| }{inst} & sat &bpdec sat&bpdec t&bpdec
        % it &pbpsat&pbp t& pbpiter
        \multicolumn{1}{ c| }{10} & 10 & 9 & 707.74 & 9195 &  5 & 37 & 257\\
        \cline{2-10}
        &\multicolumn{1}{ |c| }{e0ddr2} &
        \multicolumn{1}{ c| }{10} & 10 & 5 & 3640.40 & 26544 &  7 & 74.49 & 366 \\
        \cline{2-10}
        &\multicolumn{1}{ |c| }{ewddr2} &
        \multicolumn{1}{ c| }{10} & 10 & 10 & 10871.96 & 48053 &  9 & 21.24 & 72 \\
        \tabucline[1.2pt]{-}
        \multicolumn{1}{c|}{\multirow{1}{*}{Schurr's Lemma}} &
        \multicolumn{1}{ |c| }{-} &
        % \multicolumn{1}{ c| }{inst} & sat &bpdec sat&bpdec t&bpdec
        % it &pbpsat&pbp t& pbpiter
        \multicolumn{1}{ c| }{10} & N/A & 1 & 39.89 & 120152 &  2 & .97 & 100\\
        \tabucline[1.2pt]{-}
        \multicolumn{1}{c|}{\multirow{2}{*}{Ramsey}} &
        \multicolumn{1}{ |c| }{Ramsey 3} &
        % \multicolumn{1}{ c| }{inst} & sat &bpdec sat&bpdec t&bpdec
        % it &pbpsat&pbp t& pbpiter
        \multicolumn{1}{ c| }{8} & N/A & 1 & .01 & 61 &  4 & .75 & 283\\
        \cline{2-10}
        &\multicolumn{1}{ |c| }{Ramsey 4} &
        \multicolumn{1}{ c| }{8} & N/A & 2 & 12941.51 & 561300 &  7 & 7.39 & 81\\
        \tabucline[1.2pt]{-}
        \multicolumn{1}{c|}{\multirow{1}{*}{Chessboard Coloration}} &
        \multicolumn{1}{ |c| }{-} &
        % \multicolumn{1}{ c| }{inst} & sat &bpdec sat&bpdec t&bpdec
        % it &pbpsat&pbp t& pbpiter
        \multicolumn{1}{ c| }{14} & N/A & 5 & 35.51 & 3111 &  5 & .66 & 27\\
        \tabucline[1.2pt]{-}
        \multicolumn{1}{c|}{\multirow{1}{*}{Hanoi}} &
        \multicolumn{1}{ |c| }{-} &
        % \multicolumn{1}{ c| }{inst} & sat &bpdec sat&bpdec t&bpdec
        % it &pbpsat&pbp t& pbpiter
        \multicolumn{1}{ c| }{3} & 3 & 3 & .48 & 12 &  3 & .52 & 14\\
        \tabucline[1.2pt]{-}
        \multicolumn{1}{c|}{\multirow{1}{*}{Golomb Ruler}} &
        \multicolumn{1}{ |c| }{Arity 3} &
        % \multicolumn{1}{ c| }{inst} & sat &bpdec sat&bpdec t&bpdec
        % it &pbpsat&pbp t& pbpiter
        \multicolumn{1}{ c| }{8} & N/A  & 2 & 1.39 & 103 &  2 & 19.78 & 660\\
        \tabucline[1.2pt]{-}
        \multicolumn{1}{c|}{\multirow{1}{*}{Queens}} &
        \multicolumn{1}{ |c| }{queens} &
        % \multicolumn{1}{ c| }{inst} & sat &bpdec sat&bpdec t&bpdec
        % it &pbpsat&pbp t& pbpiter
        \multicolumn{1}{ c| }{8} & 8 & 7 & 3.30 & 159 &  8 & 2.43 & 57\\
        \tabucline[1.2pt]{-}
        \multicolumn{1}{c|}{\multirow{1}{*}{Multi-Knapsack}} &
        \multicolumn{1}{ |c| }{mknap} &
        % \multicolumn{1}{ c| }{inst} & sat &bpdec sat&bpdec t&bpdec
        % it &pbpsat&pbp t& pbpiter
        \multicolumn{1}{ c| }{2} & 2 & 2 & 2.44 & 6 &  2 & 4.41 & 10\\
        \tabucline[1.2pt]{-}
        \multicolumn{1}{c|}{\multirow{1}{*}{Driver}} &
        \multicolumn{1}{ |c| }{-} &
        % \multicolumn{1}{ c| }{inst} & sat &bpdec sat&bpdec t&bpdec
        % it &pbpsat&pbp t& pbpiter
        \multicolumn{1}{ c| }{7} & 7 & 5 & 10.14 & 1438 &  5 & 4.74 & 274\\
        \tabucline[1.2pt]{-}
        \multicolumn{1}{c|}{\multirow{1}{*}{Composed}} &
        \multicolumn{1}{ |c| }{25-10-20} &
        % \multicolumn{1}{ c| }{inst} & sat &bpdec sat&bpdec t&bpdec
        % it &pbpsat&pbp t& pbpiter
        \multicolumn{1}{ c| }{10} & 10 & 8 & 1.62 & 695 &  5 & .17 & 38\\
        \tabucline[1.2pt]{-}
        \multicolumn{1}{c|}{\multirow{2}{*}{Langford}} &
        \multicolumn{1}{ |c| }{lagford-ext} &
        % \multicolumn{1}{ c| }{inst} & sat &bpdec sat&bpdec t&bpdec
        % it &pbpsat&pbp t& pbpiter
        \multicolumn{1}{ c| }{4} & 2 & 0 & N/A & N/A &  1 & .002 & 10\\
        \cline{2-10}
        &\multicolumn{1}{ |c| }{lagford 2} &
        \multicolumn{1}{ c| }{22} & N/A & 4 & .67 & 127 &  10 & 11.64 & 10\\
        \cline{2-10}
        &\multicolumn{1}{ |c| }{lagford 3} &
        \multicolumn{1}{ c| }{20} & N/A & 0 & N/A & N/A & N/A & N/A & N/A\\
        \tabucline[2pt]{-}
      \end{tabu}
    }
  \end{center}
\end{table}

\end{document}